\documentclass[conference]{IEEEtran}
\IEEEoverridecommandlockouts
\usepackage{amsmath,amssymb,amsfonts}
\usepackage{algorithmic}
\usepackage{graphicx}
\usepackage{textcomp}
\usepackage{xcolor}
\usepackage{graphics}
\usepackage{subfigure}
\usepackage{tabularray}
\usepackage{booktabs}
\usepackage[numbers]{natbib}
\usepackage{url}

\makeatletter
\newcommand{\linebreakand}{%
  \end{@IEEEauthorhalign}
  \hfill\mbox{}\par
  \mbox{}\hfill\begin{@IEEEauthorhalign}
}
\makeatother

\def\BibTeX{{\rm B\kern-.05em{\sc i\kern-.025em b}\kern-.08em
    T\kern-.1667em\lower.7ex\hbox{E}\kern-.125emX}}
    
\begin{document}

\title{Vision Language Model Fusion for Explainable Face Recognition\\
\thanks{$^*$Corresponding author.}
}

\author{\IEEEauthorblockN{Ana Estrada-Real$^*$,
Lydia Alapatt,
Christoph Busch,
Christian Rathgeb}\\
\IEEEauthorblockA{da/sec Biometrics and Security Research Group\\Hochschule Darmstadt, Germany\\
\texttt{\{ana.estrada-real,christian.rathgeb\}h-da.de}
}}


\maketitle

\begin{abstract}
Responsible deployment of face verification systems requires more than accurate decisions: systems should also provide interpretable and auditable evidence that enables users to understand, assess, and challenge their decisions. Vision-language models (VLMs) provide a promising foundation for explainable face recognition by combining visual analysis with natural-language reasoning. However, relying on a single model may further limit the decision accuracy as well as provided explanations. This work therefore investigates whether multiple VLMs can be combined not only to improve recognition accuracy, but also to enrich the explanations associated with those decisions. To this end, this work evaluates four VLMs  as standalone face verification systems and subsequently proposes a decision-fusion framework, where two source models provide similarity scores and textual justifications and a third VLM acts as a decider model. Four different (multimodal) fusion scenarios are considered, progressively providing the decider model with scores, justifications, face images, and combinations of these modalities.

Overall, the findings suggest that the value of multi-VLM fusion extends beyond recognition performance. Different VLMs can provide complementary justifications and perspectives that enable richer explanations of face recognition decisions, supporting greater transparency, auditability, and error analysis. This is particularly relevant to the development of responsible explainable face verification systems, where users and operators should be able to understand not only the final decision but also the evidence and potential sources underlying it. The proposed multimodal VLM, which combines decision scores, explanations, and face images, achieves higher recognition accuracy than state-of-the-art VLMs and domain-specific face recognition models, while also providing fused explanations that are expected to be more robust than those generated by individual VLMs.
\end{abstract}


\section{Introduction}

Biometric recognition systems have become an increasingly important component of digital identity, authentication, border management, law enforcement, financial services, and access-control infrastructures. Among biometric modalities, face recognition (FR) is particularly attractive because it can operate passively and at a distance, without requiring explicit physical interaction from the subject. Over the past decade, deep learning has transformed FR from a relatively constrained pattern-recognition problem into a highly capable large-scale biometric technology \cite{Taigman-DeepFace-CVPR-2014, Schroff-FaceNet-CVPR-2015, Liu_FR_CVPR_1017, Wang_FR_CVPR_2018, Deng_ArcFace_CVPR_2019, Meng_MagFace_CVPR_2021}. However, the ubiquitous  deployment of FR also introduces substantial risks to privacy, fairness, security, and fundamental rights \cite{Floridi_AIEthics_MaM_2018, Jobin_AIEthics_NMI_2019, Morley_AIEthics_SEE_2019, Mehrabi_Fairness_ACMC_2021, Crawford_AIEthics_YalePress_2021}. Consequently, the development of FR systems should not be evaluated solely according to recognition accuracy: responsible deployment requires systems whose behaviour can be understood, scrutinized, challenged, and appropriately overseen by humans \cite{Baker_XAI_Corr_2023, Gebru_Data_ACM_2021, Suresh_ML_ACM_2021}.

The need for such responsibility is particularly evident in FR, where impressive empirical performance can create an unwarranted perception of infallibility. Modern FR models such as CosFace \cite{Wang_FR_CVPR_2018}, ArcFace \cite{Deng_ArcFace_CVPR_2019}, MagFace \cite{Meng_MagFace_CVPR_2021}, and AdaFace \cite{Kim_AdaFace_CVPR_2022} have achieved remarkable performance by learning highly discriminative representations of faces. These advances have substantially reduced recognition errors on established benchmarks and enabled robust recognition under challenging variations in pose, illumination, expression, and image quality \cite{Liu_FR_CVPR_1017, Wang_FR_CVPR_2018, Deng_ArcFace_CVPR_2019, Meng_MagFace_CVPR_2021, Kim_AdaFace_CVPR_2022}. Nevertheless, high accuracy does not imply uniform reliability across populations, acquisition conditions, or operational scenarios. Large-scale evaluations by NIST have demonstrated demographic differentials in both false-positive and false-negative rates, while earlier work has shown that estimates of recognition performance can depend strongly on the demographic composition of the comparison population \cite{Grother_FR_NIST_2019, Grother_Identification_NIST_2019, OToole_FRDemographcis_IVC_2012, Buolamwini_GenderShades_PMLC_2018}. Moreover, the increasing complexity of modern deep models makes it difficult for operators to understand why a particular pairwise comparison of face images led to acceptance or rejection. This creates a potentially problematic situation in which humans may defer to a highly accurate system without being able to independently assess whether a particular decision is justified, as it has been documented in several cases in which the wrong individual has been identified as a criminal by relying on FR technology alone \cite{Benedict_LawFR_WLLaw_2022}.

Concepts of explainable FR have therefore been proposed not merely as a means of making black-box models more understandable, but as a mechanism for supporting human oversight \cite{Ribeiro_AI_CoRR_2016, Lundberg_InterpretableAI_CoRR_2017, Selvaraju_GradCAM_ICCV_2017, Zhou_DL_CoRR_2015, DoshiVelez_TowardsAR_arXiv_2017}. Classical explainable-AI approaches such as LIME \cite{Ribeiro_AI_CoRR_2016}, SHAP \cite{Lundberg_InterpretableAI_CoRR_2017}, and gradient- or activation-based visualization methods \cite{Selvaraju_GradCAM_ICCV_2017, Zhou_DL_CoRR_2015} have established important foundations for interpreting complex models. However, the existence of an explanation does not necessarily imply that the explanation faithfully represents the model's actual decision process \cite{DoshiVelez_TowardsAR_arXiv_2017, Rudin_XAI_NATMI_2019, Adebayo_XAI_CoRR_2018}. This distinction is particularly important for biometric applications: an explanation that is visually compelling but causally unrelated to the recognition decision could reduce, rather than increase, human trust.

The regulatory landscape further strengthens the need for responsible and explainable FR. The European Union's Artificial Intelligence Act (AI Act), Regulation (EU) 2024/1689, explicitly places transparency, human oversight, accuracy, robustness, and risk management among the requirements for high-risk AI systems (including FR) \cite{EU-Regulation-AI-2024-1689-en-240712}. In particular, Article 13 requires high-risk systems to provide sufficient transparency to enable operators to interpret outputs and use systems appropriately, including information about capabilities, limitations, accuracy, risks, and, where applicable, mechanisms that provide information relevant to explaining outputs \cite{EUAIAct_Article13_2024}. Article 14 further requires effective human oversight and explicitly recognizes the risk of over-reliance on AI outputs (automation bias); for certain remote biometric identification applications, additional verification by competent humans is required \cite{EUAIAct_Article14_2024}.

The research community has responded to these challenges by investigating several complementary forms of explainable FR. Early approaches primarily explored spatial explanations, identifying facial regions that contribute to a verification decision. Recent work has proposed feature-guided gradient back-propagation to produce similarity and dissimilarity maps for accepted and rejected comparisons \cite{Lu_XFR_FG_2024}, model-agnostic saliency methods such as CorrRISE  \cite{Lu_XFR_WACV_2024}, and more localized CAM-based explanations \cite{Shadman_XFR_CoRR_2025}. Other work has explored explanations in the frequency domain, arguing that spatial heatmaps can overlook frequency components that influence deep FR models \cite{Huber_XFR_WACV_2025}. Human-centered approaches have additionally investigated the correspondence between machine explanations and semantically meaningful facial regions \cite{Doh_XFR_CoRR_2024, DeAndresTame_XFR_Corr_2024a, Cascone_XFR_Elsevier_2023}. These efforts represent important progress towards making FR decisions inspectable; however, they primarily answer the question of where a model obtains evidence rather than whether that evidence constitutes a reliable, faithful, and human-actionable justification for the decision.

\subsection{Related Work}
A complementary direction  towards explainable FR has recently emerged from vision-language models (VLMs) and multimodal large language models (MLLMs), which offer the possibility of transforming visual evidence into natural-language explanations. Rather than presenting users solely with saliency maps or numerical similarity scores, VLM-based systems can describe similarities and differences between two faces in a form that is potentially more accessible to non-expert users. VerLM, for example, explicitly investigates FR together with natural-language explanations \cite{Hannan_XFRNL_CoRR_2026}, while recent work has examined MLLMs for textual explanations of unconstrained face comparisons \cite{Sony_MLLMFR_IWBF_2026}. 
These developments are particularly promising because natural-language reasoning may provide a common interface through which machine evidence can be communicated both to human operators and to other AI systems.

Table~\ref{tab:relatedwork} provides an overview of most relevant works exploring the use of VLMs for FR.
An early study by DeAndres-Tame et al. \cite{DeAndresTame_FR_Corr_2024}, evaluated ChatGPT on a range of face analysis tasks, including FR, facial expression recognition, and face description. ChatGPT achieved 93.50\% accuracy on the Labeled Faces in the Wild (LFW) dataset. Shahreza and Marcel \cite{Shahreza_FULLM_ICCVW_2025} introduced FairFaceGPT, a benchmark designed to evaluate VLMs across a broader set of face-understanding tasks, including FR, face localization, and expression recognition. The results showed that even large commercial VLMs, such as GeminiPro 1.5, achieved at most 70.0\% accuracy across the evaluated tasks. Sony et al \cite{Sony_FRLLM_ICCVW_2025} compared VLMs with state-of-the-art domain-specific FR models, such as AdaFace \cite{Kim_AdaFace_CVPR_2022}, and investigated the effect of image resolution and cropping on recognition performance. Using LFW images at various resolutions, the authors showed that VLM performance improves when less aggressively cropped images are provided, suggesting that VLMs can benefit from contextual cues. As best VLM, OpenCLIP-H-14 achieved 81.73\%TMR@0.01\%FMR, surpassing the reported performance of AdaFace (77.31\%TMR@0.01\%FMR). More recently, Narayan et al. \cite{Narayan_FU_IEEETBIOM_2026} introduced FaceXBench, a comprehensive evaluation methodology covering a wide range of face-understanding tasks and datasets, further confirming the challenges observed in previous studies; among the evaluated VLMs, GeminiPro 1.5 achieved the highest performance on high resolution FR tasks. Finally, one of the most recent studies of Shahreza and Marcel \cite{Shahreza_FRLLM_ICASSP_2026}, reports a substantial improvement in VLM-based FR, with the open-source Qwen2-VL-7B-Instruct achieving 93.28\% accuracy on LFW, approaching the performance that ChatGPT had previously and narrowing the gap with specialized FR models. Overall, these studies indicate a rapid progression in VLM-based FR, while still highlighting the persistent performance gap between VLMs and domain specific systems.

\begin{table*}
\centering
  \caption{Best performing VLMs evaluated for FR in recent benchmarks.}
  \label{tab:relatedwork}
  \begin{tabular}{lrrrrr}
    \toprule
    Year & Reference & Top VLM & Datasets tested & Performance \\
    \midrule
    2024 & DeAndres-Tame et al. \cite{DeAndresTame_FR_Corr_2024} & ChatGPT1x1 & LFW & 93.50\% Acc. \\
    2025 & Shahreza and Marcel \cite{Shahreza_FULLM_ICCVW_2025}& GeminiPro1.5 & Various (8 Mixed) & 70.00\% Acc. \\
    2025 & Sony et al. \cite{Sony_FRLLM_ICCVW_2025} & OpenCLIP-H-14 & LFW & 81.73\% TMR@0.01\%FMR \\
    2026 & Narayan et al. \cite{Narayan_FU_IEEETBIOM_2026} & GeminiPro 1.5 & Various (26 Mixed) & 82.25\% Acc.  \\
    2026 & Shahreza and Marcel \cite{Shahreza_FRLLM_ICASSP_2026} & Qwen2-VL-7B-Instruct & LFW & 93.28\% Acc. \\
  \bottomrule
\end{tabular}
\end{table*}

The use of VLMs is motivated not only by their computer vision capabilities, but also to the access to natural language descriptions that can be used to understand FR scores. Some of the previous works \cite{DeAndresTame_FR_Corr_2024, Sony_FRLLM_ICCVW_2025} show that MLLMs can produce textual reasoning while comparing two face images even in ambiguous scenarios. This has also been further analysed \cite{Sony_MLLMFR_IWBF_2026} giving VLMs auxiliary FR information (scores and decisions), show improvement in FR performance while simultaneously referring to non-verifiable or hallucinated facial attributes. To work on this issue, a dedicated VLM, VerLM \cite{Hannan_XFRNL_CoRR_2026} was trained specifically on faces (VGGFace2) to produce descriptions for image pairs. These descriptions seem to stay consistent to the aligned face embeddings of the image pairs, but ground truth descriptions were generated by a Llama2 and no hallucination analysis is presented.

Sony et al. \cite{Sony_FRLLM_ICCVW_2025} further showcased that the fusion of traditional FR model and VLMs can lead to improved FR performance. In their work, a trivial fusion is performed on score level, which is common practise for information fusion in biometric systems including FR. In contrast, a multimodal fusion of VLMs incorporating scores, explanations and images has not yet be investigated for the task of FR (to the best of the authors' knowledge).

\subsection{Contribution and Organization}

This work investigates the use of VLMs not only as FR systems capable of producing explanations, but also as reasoning agents that can inform and constrain the decisions of other VLMs. The central premise is that the reasoning capabilities of multiple VLMs can be leveraged to (a) provide complementary assessments of facial comparisons, (b) identify inconsistencies between model outputs and explanations, and (c) reduce the dependence of the final explanation on the potentially unreliable linguistic priors of a single model. In this setting, explanations become a component of a broader recognition pipeline. Specifically, the proposed multimodal fusion framework produces a biometric decision, exposes the evidence supporting that decision, and communicates this information to a downstream VLM for further reasoning. This formulation moves beyond conventional post-hoc visualization techniques towards a framework in which recognition, explanation, cross-model reasoning, and human oversight are considered jointly.

This work makes the following contributions:
\begin{itemize}
\item A comprehensive FR benchmark of state-of-the-art VLMs on an established FR dataset using standardized performance metrics.
\item Proposal of various fusion strategies, including multimodal fusion together with a detailed analysis of the fusion methods, identifying the best configurations that outperform standalone VLMs as well as dedicated FR models.
\item A qualitative analysis of the explanations provided by the VLMs and their evolution before and after the proposed fusion.
\end{itemize}

The remainder of this paper is organized as follows. Section~\ref{sec:Methodology} describes the dataset, baseline FR algorithms, selected VLMs, fusion scenarios, and qualitative evaluation of the generated explanations. Section~\ref{sec:Results} presents the results for the baseline FR algorithms, standalone VLMs, fusion scenarios, and explanation analysis. These results, their improvements and limitations are then discussed in Section \ref{sec:Discussion}. Finally, Section~\ref{sec:Conclusions} presents the main conclusions and discusses directions for future work.

\section{Methodology}
\label{sec:Methodology}

\subsection{Dataset}

The Labeled Faces in the Wild (LFW) dataset \cite{Huang_LFWTech_2007} was selected as the primary evaluation dataset for this study. LFW contains more than 13,233 facial images belonging to 5,749 individuals and was specifically designed to represent FR under unconstrained conditions. The images in LFW exhibit substantial variation in factors such as pose, illumination, facial expression, image quality, background, and appearance.

The use of unconstrained facial images is particularly relevant to the objectives of this work. Since the proposed approach investigates not only FR performance but also the quality and reliability of model-generated explanations, evaluating the models under relatively challenging and diverse visual conditions provides a more demanding test of their multimodal reasoning capabilities. An explanation that appears plausible for two high-quality, similarly aligned facial images may be substantially less reliable when the images differ in pose, illumination, expression, or image quality. Therefore, the variability present in LFW provides an appropriate environment for investigating whether VLM-generated explanations remain meaningful and consistent under unconstrained conditions.

For the FR experiments, image pairs were compiled from the LFW dataset. A total of 200,000 image pairs were generated, consisting of an equal proportion of mated and non-matd pairs (50:50). This balanced construction ensures that neither class dominates the evaluation and allows the performance of the models to be assessed consistently across both positive and negative FR decisions.

Each pair is associated with a binary ground-truth label indicating whether the two images represent the same identity. These labels are used exclusively for the quantitative evaluation of the model outputs and are not provided to the VLMs during inference. For every pair, the models receive only the two facial images and are required to independently estimate their similarity and provide a textual justification for their decision. The same set of image pairs is subsequently used across the evaluated models and fusion configurations, ensuring that differences in performance can be attributed to the model and fusion strategy rather than differences in the evaluated image pairs.

\subsection{Dedicated FR Baselines}

To contextualize the FR performance of the VLM-based approaches, the same LFW-derived evaluation pairs are also evaluated using established dedicated FR models. Specifically, AdaFace \cite{Kim_AdaFace_CVPR_2022}, LVFace \cite{You_LVFace_ICCV_2025}, and MagFace \cite{Meng_MagFace_CVPR_2021} are used as reference baselines. These models are designed specifically for FR and are therefore expected to provide a strong comparison against the more general-purpose VLMs investigated in this work. Note that the dedicated FR algorithms require face images to undergo preprocessing. For this purpose, AdaFace's face alignment algorithm was used. Faces were detected based on five facial landmarks and subsequently cropped to a resolution of 112×112×3 pixels. In contrast, recent work examining performance differences between dedicated FR algorithms and VLMs found that VLM performance on zero-shot FR tasks improved when using non-cropped images, suggesting that VLMs may benefit from contextual cues \cite{Sony_FRLLM_ICCVW_2025}. Therefore, the face images used for the VLMs were retained at their original resolution of 250×250×3 pixels.

For consistency, the three baseline models are evaluated on the same image pairs used in the VLM experiments. Their outputs are used to calculate the same FR metrics, including Equal Error Rate (EER), the corresponding EER threshold, FNMR at 0.1\% FMR, and FNMR at 0.01\% FMR. This ensures that the performance of the VLM-based systems and dedicated FR models is evaluated under identical image-pair conditions and operating points.

The baseline comparison serves two purposes. First, it provides a reference for determining how the recognition performance of general-purpose VLMs compares with that of models specifically developed for FR. Second, it allows the improvements obtained through VLM fusion to be interpreted relative to established FR systems. The objective is not to expect VLMs to necessarily outperform specialized FR models, but rather to determine whether multimodal models can achieve competitive FR performance while providing an additional capability in the form of natural-language explanations.

\subsection{Vision Language Models}

A set of 15 available open source VLMs were initially selected from Hugging-Face's library for pre-test, they were given 3,200 image pairs to evaluate and the best 4 were selected for the full dataset evaluation. As shown in Table \ref{tab:models}, the models Gemma4-31B-it, InternVL3-8B, Ovis-2-6-30B-A3B, and Qwen3-VL-8B-Instruct were selected to provide variation in model architecture and scale, while maintaining a common multimodal interface for the FR task. In particular, two models operate at approximately 8 billion parameters, while the remaining two contain approximately 30 billion parameters, as specified in Table \ref{tab:models}. This selection enables the investigation of whether model scale is associated with differences in FR accuracy, explanation quality, or the ability to benefit from the responses of other VLMs during the fusion stage.

The models were obtained from the Hugging Face model repository and executed locally using a graphics card NVIDIA A100 80GB. To ensure reproducibility and minimize variations arising from stochastic generation, all models were evaluated with a temperature of 0, a fixed random seed of 42, a maximum token limit of 200 for the standalone experiments, 30 tokens for the first and second fusion and 275 for the third and fourth fusion.  These quantities are directly related to the amount of information that is provided to the models. 


\begin{table}
  \caption{Open source VLMs used for this work.}
  \label{tab:models}
  \begin{tabular}{lrrrrr}
    \toprule
    Model & Parameters & Max. context & Size (GB) \\
    \midrule
    Gemma4-31B-it \cite{ElAbd_Gemma4_arXiv_2026} & 31B & 256k & 62.6 \\
    InternVL3-8B \cite{Zhu_Intern_arXiv_2025} & 8B & 40k & 15.9\\
    Ovis-2.6-30B-A3B \cite{Lu_Ovis_arXiv_2025} & 30B & 64K & 62.8\\
    Qwen3-VL-8B-Instruct \cite{Bai_Qwen_arXiv_2025} & 8B & 256K & 17.5\\
  \bottomrule
\end{tabular}
\end{table}

\begin{figure}[h]
  \centering
 \includegraphics[width=0.95\linewidth, alt={Two icons with human silhuetes pointing at an icon representing a vision language model then pointing to an results sheet.}]{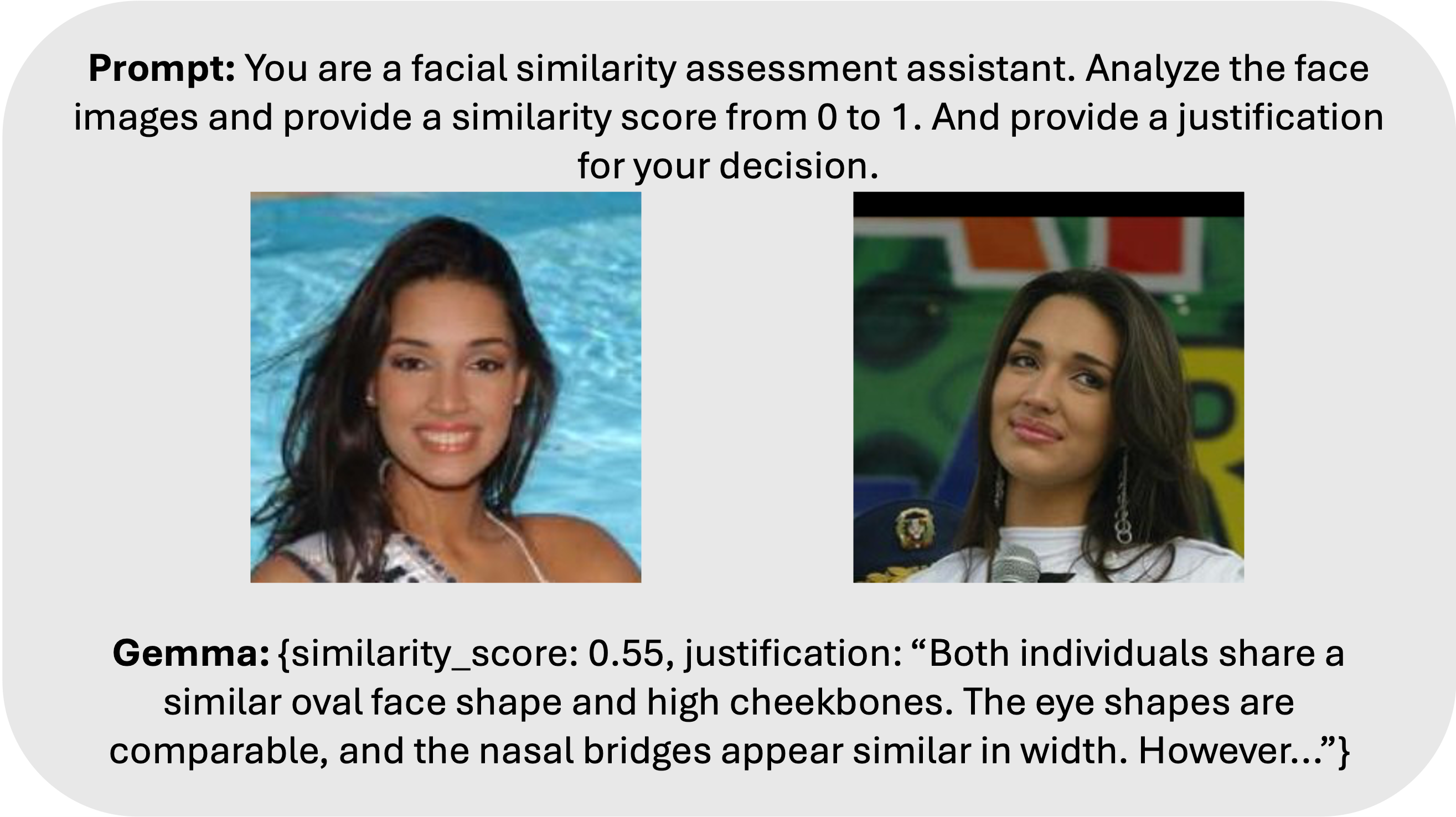}
  \caption{Diagram representing the models' standalone FR test. Each model would get the 200,000 image pairs and be asked to output a similarity score between $[0,1]$ and a justification for each of them.}
  \label{fig:prompt}
\end{figure}

For each evaluation pair, the VLM receives two facial images as visual input and a prompt. The models are instructed to perform a zero-shot FR assessment by estimating the degree of similarity between the two faces present in the images. Each model is required to produce two outputs: (1) a numerical similarity score in the range [0,1], and (2) a textual justification explaining the basis for its assessment with a requirement to maximum 200 tokens, as shown in Figure \ref{fig:prompt}. A score closer to 1 represents a higher degree of perceived facial similarity, whereas a score closer to 0 represents lower similarity. The textual justification is intended to describe the visual characteristics that the model considers relevant to its decision, the models were recommended to look at facial attributes such as eyebrows, eyes, nose, mouth, chin, jawline, cheekbones, and to ignore background and clothing.

The outputs of each VLM are stored together with the corresponding image-pair identifier and ground-truth label. The numerical similarity scores are subsequently used for the standalone FR evaluation, while both the scores and textual justifications are retained as inputs for the subsequent fusion experiments. Importantly, the ground-truth labels are used only for quantitative evaluation and are not provided to the VLMs during either standalone inference or fusion. Consequently, every model must independently derive its similarity assessment and explanation from the available visual and model-generated evidence.

The resulting experimental design provides two complementary comparisons. First, the four VLMs can be evaluated independently to determine whether differences in model scale and architecture affect FR performance and explanation quality. Second, their stored responses can be combined in the proposed (multimodal) fusion framework to investigate whether a VLM can benefit from the complementary predictions and explanations generated by other models.

\subsection{Fusion}

Following the standalone evaluation, a second experimental stage investigates whether the outputs of multiple VLMs can be combined to improve FR performance and explanation quality. The fusion strategy uses two VLMs as source models and a third, distinct VLM as the decider model. The two source models independently process the same facial image pair and produce their similarity scores and textual justifications. The outputs of these two models are subsequently provided to a third VLM, which is responsible for producing the final similarity score and justification resulting in 12 unique source-pair/decision-model combinations. Each of these 12 combinations is evaluated under four different information-sharing scenarios, resulting in a total of 48 fusion configurations. The four scenarios are designed to progressively increase the amount of information available to the decider model and to isolate the contribution of numerical predictions, textual explanations, and the original visual evidence.

\begin{figure*}
  \centering
 \includegraphics[width=0.90\linewidth, alt=Four icons representing four VLMs as standalone models pointing at the four different scenarios each of them feeding a decider VML which then outputs new results.]{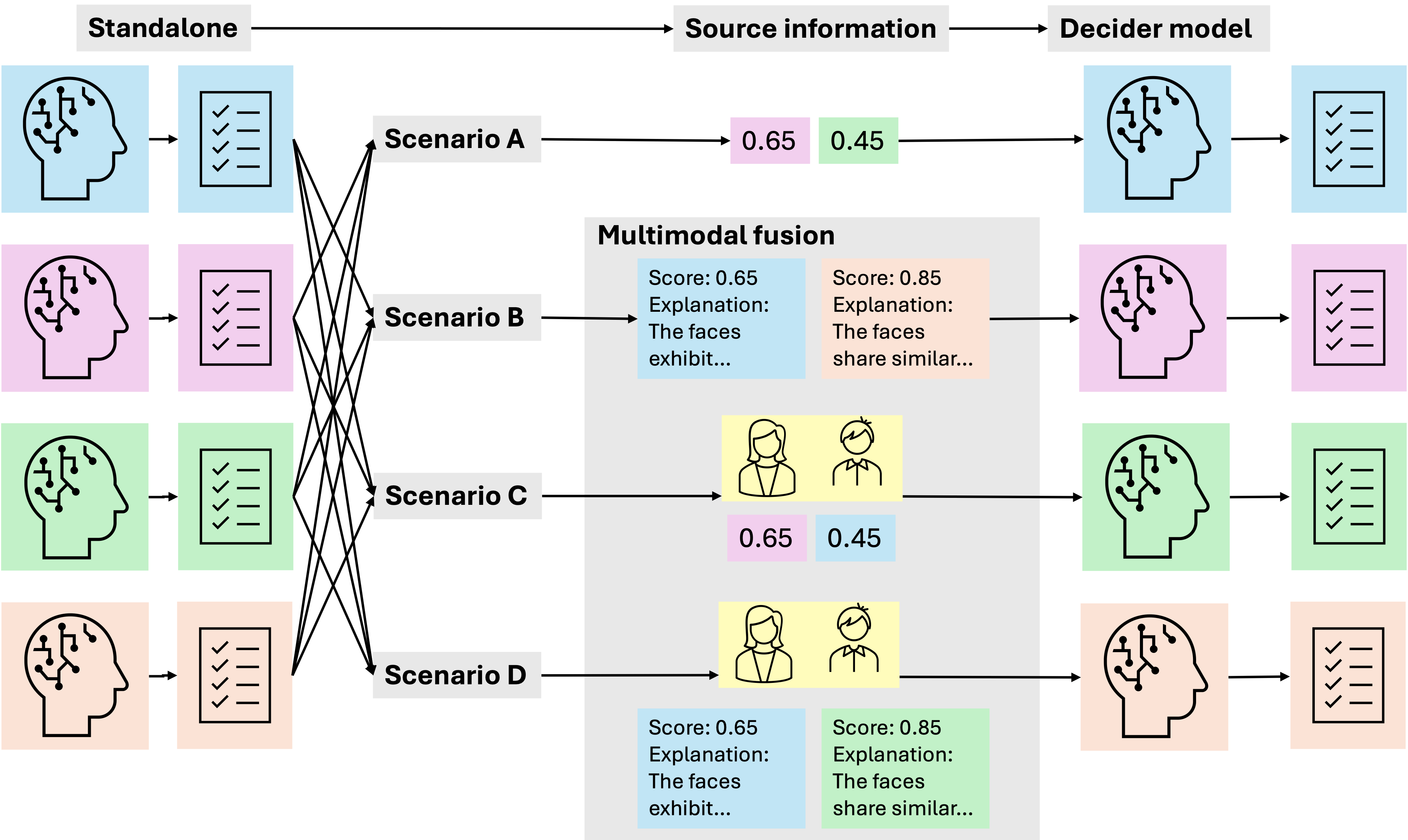}
  \caption{Diagram of the models' fusion. Each model would produce a similarity score and justification for each image pair. This information will then be used part of the fusion. There are four scenarios: scenario A where the decider model will get the similarity scores of two source models, scenario B where the decider model will get the similarity scores and justification of two source models. For scenarios C and D there heterogeneous data-sources are used, scenario C where the decider model has access to the original face image pairs and the similarity scores of two source models, and scenario D where the decider model will also have access to the original face image pairs, in addition to the similarity scores and justifications.}
  \label{fusion}
\end{figure*}

The four proposed fusion scenarios are defined as follows:

\begin{enumerate}
    \item \textbf{Scenario A: Similarity scores.} The decider model receives only the similarity scores produced by the two source models. The original facial images $I_1$, $I_2$ and the textual justifications generated by the source models are not provided. This configuration evaluates whether the decider model can benefit from the numerical predictions of other VLMs alone.
    \begin{equation} (S_1,S_2) \xrightarrow{\mathrm{VLM}_3} (S_f,E_f) \end{equation}

    where $(S_1)$ and $(S_2)$ are the similarity scores generated by the two source models, i.e. $S_1={VLM}_1(I_1,I_2)$ and $S_2={VLM}_2(I_1,I_2)$, and $(S_f)$ and $(E_f)$ represent the final similarity score and explanation produced by the decidermodel. Note that this fusion scenario reflects a traditional score level fusion.

    \item \textbf{Scenario B: Similarity scores and justification.} The decider model receives the similarity scores and textual justifications generated by both source models, i.e.  i.e. $S_1, E_1={VLM}_1(I_1,I_2)$ and $S_2, E_2={VLM}_2(I_1,I_2)$,. The original facial images are not provided during the fusion stage. This configuration evaluates whether the semantic information contained in the source explanations provides complementary information beyond the numerical scores.
    \begin{equation} (S_1,E_1,S_2,E_2) \xrightarrow{\mathrm{VLM}_3} (S_f,E_f) \end{equation}

    \item \textbf{Scenario C: Image pair and similarity scores.}The decider model receives the original pair of facial images together with the similarity scores produced by the two source models. This configuration allows the decider model to independently inspect the visual evidence while also considering the numerical predictions of the source models.
    \begin{equation} (I_1,I_2,S_1,S_2) \xrightarrow{\mathrm{VLM}_3} (S_f,E_f) \end{equation}

    \item \textbf{Scenario D: Image pair, similarity score and justification.}The decider model receives the original facial images, the similarity scores, and the textual justifications produced by both source models. This represents the full-information fusion configuration and allows the decider model to jointly consider the original visual evidence and the outputs of the other VLMs.
    \begin{equation} (I_1,I_2,S_1,E_1,S_2,E_2) \xrightarrow{\mathrm{VLM}_3} (S_f,E_f)
    \end{equation}
\end{enumerate}

The four configurations therefore provide a controlled progression in the information available to the decider model, as summarized in Figure \ref{fusion}. Scenarios B to D represent multimodal fusion scenarios in which a final decision is obtained from the fusion of heterogeneous inputs.

For every fusion configuration, the decider model is instructed to produce a new similarity score in the range ([0,1]) and only in scenarios C and D a new justification will be generated. The final scores are evaluated using the same FR metrics as the standalone models, Equal Error Rate (EER), the corresponding EER threshold, FNMR at 0.1\% FMR, and FNMR at 0.01\% FMR. This enables a direct comparison between standalone and fused predictions.

Overall, the fusion experiments address two principal research questions: first, if the information generated by multiple VLMs improve the accuracy and robustness of FR compared to standalone VLM inference; second, if access to complementary model predictions, textual justifications, and visual evidence improve the quality and reliability of the final explanation.

\subsection{Explanation Quality Evaluation}

In addition to the quantitative evaluation of FR performance, a case-based qualitative analysis is conducted to investigate whether fusion affects the quality and reliability of the explanations generated by the VLMs. Rather than evaluating explanations only on correctly classified samples, the analysis considers cases representing different levels and types of model uncertainty. This is important because an explanation may appear convincing for an unambiguous correct decision while being less informative or even misleading when the model is uncertain or produces an incorrect decision.

Representative edge cases are selected from the standalone results for each of the four VLMs. The same image pairs are subsequently evaluated again during the fusion experiments, allowing the standalone and fused explanations to be compared on identical visual inputs. The cases are selected according to the model's similarity score, predicted decision, and ground-truth identity relationship.

The three cases selected for each model are defined as follows:

\begin{enumerate}
    \item \textbf{Correct near-threshold case:} a correctly classified image pair whose similarity score is closest to the model's EER threshold. This case represents a relatively ambiguous decision near the model's operating boundary and is used to examine how the model explains a difficult but ultimately correct decision.
    \item \textbf{High-confidence false negative:} an image pair, consisting of two images of the same individual, that is incorrectly classified as non-mated, i.e. non-matching, with the highest degree of confidence. This case is used to investigate whether the explanation identifies visual factors that may have contributed to the erroneous rejection and whether the model appropriately communicates the difficulty of the comparison.
    \item \textbf{High-confidence false positive:} an pair, consisting of images of different individuals, that is incorrectly classified as mated, i.e. matching, with the highest degree of confidence. This case is used to examine whether the explanation identifies the visual similarities that may have led to the erroneous match and whether the model expresses unjustified certainty in its reasoning.
\end{enumerate}

The selection criteria are determined directly from the model's similarity scores and EER threshold. For a correctly classified near-threshold case, the selected pair minimizes the absolute difference between the predicted score and the EER threshold: \begin{equation} i^* = \arg\min_i |s_i-t_{\mathrm{EER}}|\end{equation}

where $(s_i)$ is the similarity score and $(t_{\mathrm{EER}})$ is the model-specific EER threshold. For false negatives, the case with the lowest similarity score among genuine pairs classified as non-matches is selected. Conversely, for false positives, the case with the highest similarity score among impostor pairs classified as matches is selected. This procedure provides an objective and reproducible definition of the selected edge cases.

The explanation analysis is performed through direct qualitative inspection rather than through an automated language-based evaluation method. In particular, methods such as BERT-based semantic similarity measures or LLM-as-a-Judge approaches are not used to assign an automatic explanation score. This choice is motivated by the absence of a ground-truth explanation for each face pair. Unlike the binary identity label used to evaluate FR performance, there is no unique reference explanation that specifies the facial characteristics that a model should identify. Consequently, an automated comparison against a reference text could penalize valid explanations expressed using different wording or, conversely, assign a high score to a fluent explanation that is not actually supported by the visual evidence.

The qualitative assessment therefore focuses on the following dimensions:

\begin{itemize}
    \item \textbf{Groundedness}: the explanation refer to visible facial properties rather than irrelevant image content.
    \item \textbf{Specificity}: identifies concrete attributes (eyes, nose, jawline, facial structure).
    \item \textbf{Discriminativity}: actually compares the two images, rather than describing each one independently.
    \item \textbf{Calibration}: the language appropriately reflect ambiguity, particularly for near-threshold examples.
    \item \textbf{Hallucinations}: the explanation mentions attributes that are not visible or irrelevant to the compared identities.
\end{itemize}

\section{Results and Discussion}
\label{sec:Results}

The experimental results are presented in a progressive manner to provide a clear comparison between dedicated FR systems, standalone VLMs, and the proposed VLM fusion approach. The selected edge cases are also revisited to compare the standalone and fused explanations and to assess how fusion affects their relevance, specificity, consistency, and error awareness. Finally, the results are discussed jointly to identify the conditions under which VLM fusion improves FR performance and explanation quality, to examine differences between model architectures and scales, and to assess the potential and limitations of VLM-based fusion for explainable FR.

\subsection{Baseline}

The three conventional FR baselines are close in overall performance. In Table \ref{tab:baseline}, MagFace achieves the best Area under Curve (AUC) value (0.9741) and lowest EER (5.57\%), making it the strongest baseline under the aggregate verification metrics. AdaFace has a slightly worse AUC/EER than MagFace, but it performs best at the very low false-match operating points: FNMR of 6.05\% at 0.1\% FMR and 6.13\% at 0.01\% FMR. LVFace is slightly behind both, with AUC 0.9704 and EER 5.78\%. MagFace provides the best overall discrimination, while AdaFace provides the strongest performance in the low-FMR region. LVFace ranks third across all evaluated metrics.

\begin{table*}
    \centering
    \caption{Baseline FR results.}
    \resizebox{\textwidth}{!}{%
    \begin{tabular}{lrrrrrrrr}
    \toprule
    Model & AUC & EER & EER Threshold & d-prime & FNMR@0.1\%FMR & Threshold@0.1\%FMR & FNMR@0.01\%FMR & Threshold@0.01\%FMR \\
    \midrule
    AdaFace \cite{Kim_AdaFace_CVPR_2022} & 0.9718 & 0.0562 & 0.0901 & \textbf{4.8732} & \textbf{0.0605} & 0.1915 & \textbf{0.0613} & 0.2343 \\
    LVFace \cite{You_LVFace_ICCV_2025} & 0.9704 & 0.0578 & 0.0836 & 4.4983 & 0.0613 & 0.1673 & 0.0621 & 0.2047 \\
    MagFace \cite{Meng_MagFace_CVPR_2021}& \textbf{0.9741} & \textbf{0.0557} & 0.1343 & 4.8431 & 0.0612 & 0.2652 & 0.0620 & 0.3201 \\
    \bottomrule
    \end{tabular}
    }
    \label{tab:baseline}
\end{table*}

\begin{figure}
  \centering
    \includegraphics[width=0.90\linewidth, alt=A plot with three DET curves repersenting the three baseline models adaFace in green lvface in orange and magface in purple.]{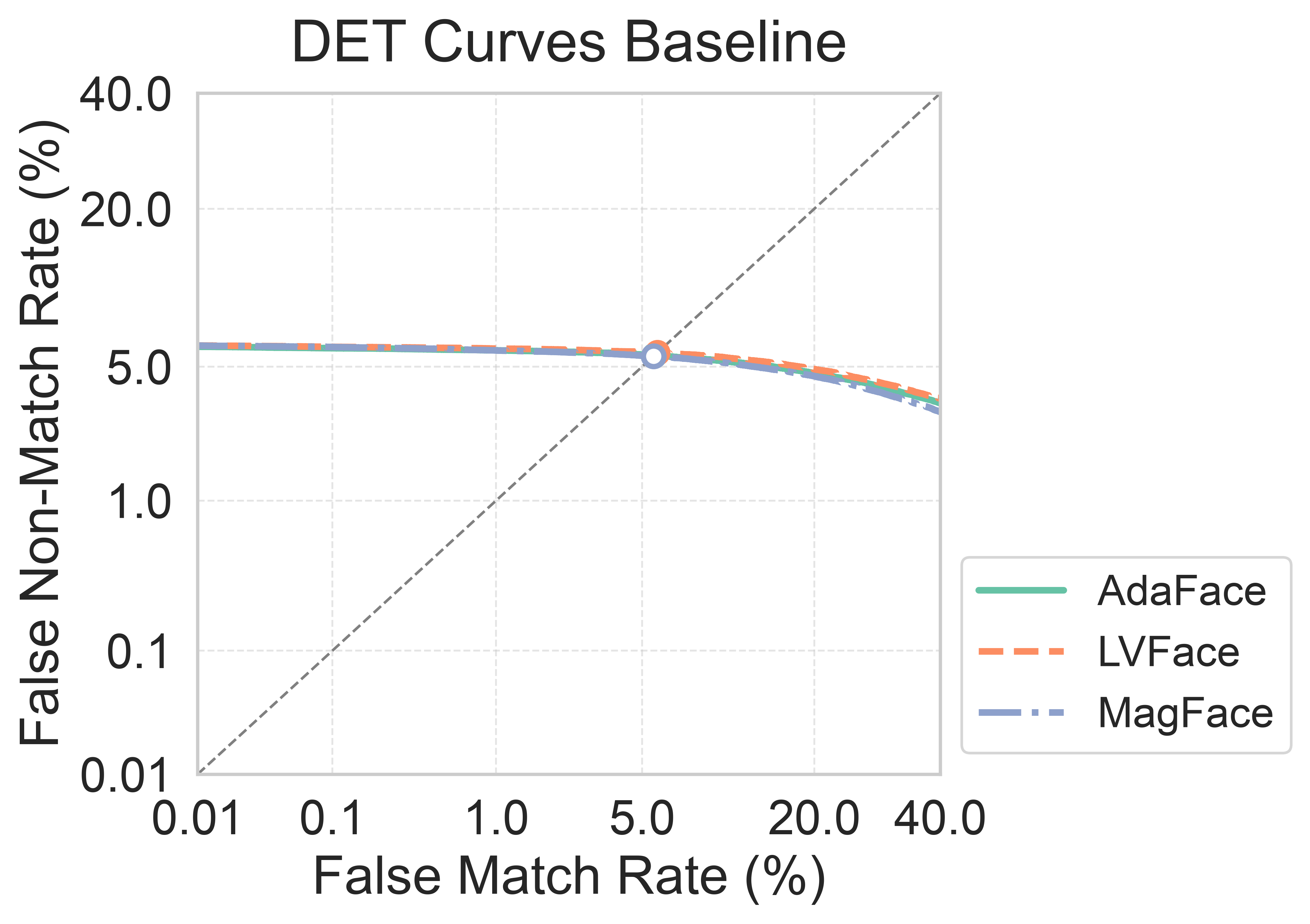}
  \caption{DET curves showing the performance of the three baseline models.}
  \label{fig:baseline}
\end{figure}

In Figure \ref{fig:baseline} the DET curves of the three baseline models exhibit a similar and smooth behaviour, indicating stable performance across the evaluated range of false match rates. All three models show a relatively flat trend from approximately 5\% FMR toward their respective EER points, followed by a gradual decrease in FNMR. MagFace achieves the lowest FNMR at around 40\% FMR, although the difference from AdaFace and LVFace is marginal. Overall, the close proximity of the curves further supports the observation that the three dedicated FR models provide broadly comparable baseline performance, with no single model exhibiting a dominant advantage across the entire operating range.

The baseline evaluation is included primarily to provide a reference point for the performance of established state-of-the-art FR methods on the selected benchmark.

\subsection{VLM Standalone}

For the standalone VLMs, Table \ref{tab:standalone} shows the evaluation along with the best performing FR model. Gemma is the strongest VLM according to most evaluation metrics. However, Gemma's excellent global metrics do not translate into the best performance at the strictest operating point. At 0.1\% FMR, its FNMR is 6.88\%, which is close to AdaFace's 6.05\%. At 0.01\% FMR, however, Gemma's FNMR rises to 21.38\%, substantially worse than the baseline models (6.1\%). On the other hand, Qwen's score distributions have very strong overall separation but do not maintain that advantage in the extreme tail of the impostor distribution. Intern and Ovis are substantially weaker than Gemma and Qwen in the global metrics. 

\begin{table*}
    \centering
    \caption{VLMs Standalone FR evaluation.}
    \resizebox{\textwidth}{!}{%
    \begin{tabular}{lrrrrrrrr}
    \toprule
    Model & AUC & EER & EER Threshold & d-prime & FNMR@0.1\%FMR & Threshold@0.1\%FMR & FNMR@0.01\%FMR & Threshold@0.01\%FMR \\ \midrule
    MagFace & 0.9741 & 0.0557 & 0.1343 & 4.8431 & 0.0612 & 0.2652 & 0.0620 & 0.3201 \\
    \midrule
    Gemma & \textbf{0.9985} & \textbf{0.0130} & 0.5488 & 6.5239 & \textbf{0.0688} & 0.8745 & \textbf{0.2138} & 0.9230 \\
    Intern & 0.9935 & 0.0329 & 0.6899 & 4.6494 & 0.3413 & 0.8653 & 0.6358 & 0.8868 \\
    Ovis & 0.9932 & 0.0299 & 0.8228 & 4.5316 & 0.3685 & 0.9138 & 0.6522 & 0.9579 \\
    Qwen & 0.9966 & 0.0152 & 0.7662 & \textbf{7.6937} & 0.1550 & 0.9508 & 0.4265 & 0.9602 \\
    \bottomrule
    \end{tabular}
    }
    \label{tab:standalone}
\end{table*}

\begin{figure}
    \centering
    \subfigure[]{\includegraphics[width=0.49\linewidth, alt=Gemma histogram nonmated in ted and mated in green.]{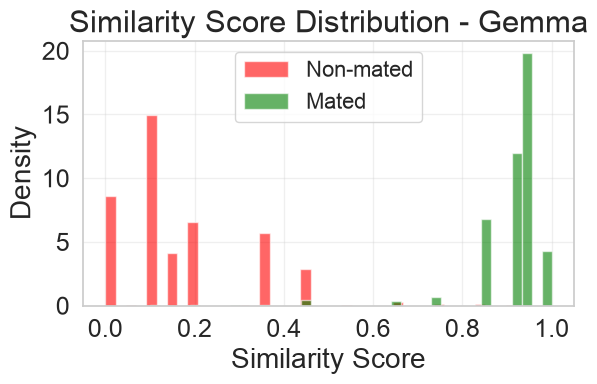}} 
    \subfigure[]{\includegraphics[width=0.49\linewidth, alt=Intern histogram nonmated in ted and mated in green.]{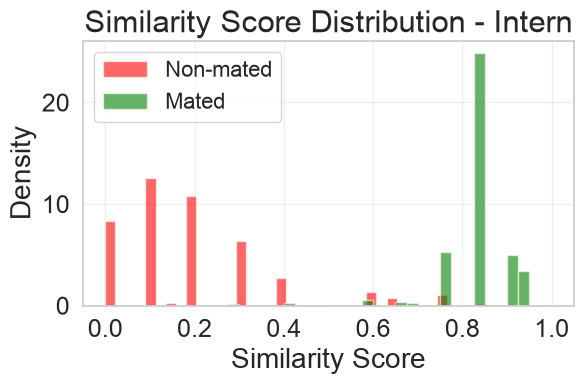}} 
    \subfigure[]{\includegraphics[width=0.49\linewidth, alt=Ovis histogram nonmated in ted and mated in green.]{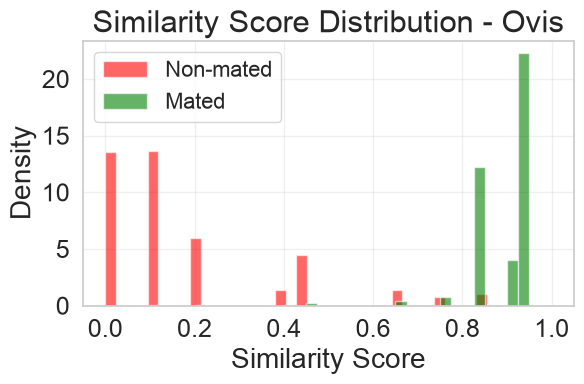}}
    \subfigure[]{\includegraphics[width=0.49\linewidth, alt=Qwen histogram nonmated in ted and mated in green.]{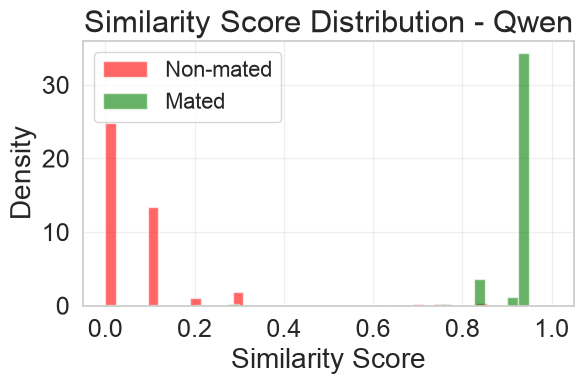}}
    \caption{Distograms of the scores distribution for (a) Gemma (b) Intern (c) Ovis and (d) Qwen.}
    \label{fig:histograms}
\end{figure}

The relatively poorer performance of the VLMs at very low FMR operating points may partly be related to the granularity of their decision scores as shown in Figure \ref{fig:histograms}. Precisely, VLMs tend to return discrete scores, mostly in steps of 0.05, i.e. 5\%. Since these operating points correspond to the extreme tail of the impostor distribution, limited score resolution can constrain the available decision thresholds and result in abrupt changes in the achievable FMR and FNMR. The observed degradation may also reflect differences in the underlying score distributions and their behaviour in the impostor tail.

\begin{figure}[h]
  \centering
 \includegraphics[width=0.90\linewidth, alt=DET curves of Gemma Intern Ovis and Qwen.]{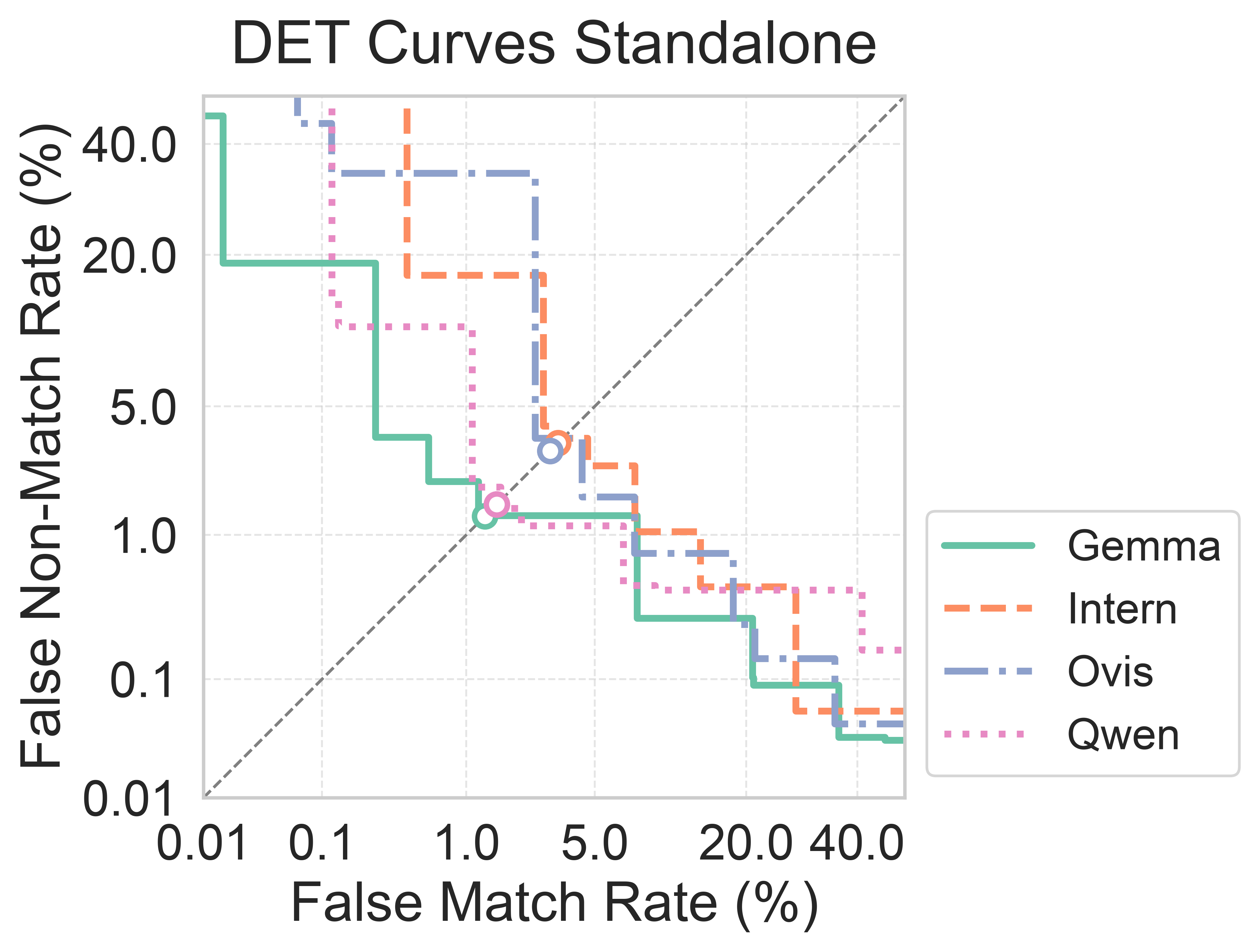}
  \caption{DET curves from the standalone VLMs.}
  \label{fig:standalone}
\end{figure}

Figure \ref{fig:standalone}, depicts the corresponding DET curves for the standalone VLMs. Note that the aforementioned discrete scores yield more steppy DET curves. Gemma has the best overall and lowest EER; particularly strong across much of the useful operating range. Qwen remains very competitive with Gemma around the EER region and at moderate FMR, but weaker at some very-low/high-FMR regions. Ovis is poor at very low FMR, but improves dramatically as FMR is relaxed. Intern is similarly poor at low FMR and generally does not beat Gemma/Qwen in the important middle region.

\subsection{VLM Fusion}

\subsubsection{Scenario A}

In scenario A the VLM got as input only the similarity scores from two source models. In Table \ref{tab:scenarioa}, the strongest configuration is Intern $\leftarrow$ Qwen + Gemma with AUC = 0.9992, EER = 1.17\%, and d-prime = 7.9473. This is the best result in Scenario A across all three of these global discrimination metrics. More importantly, its FNMR at 0.1\% FMR is 5.17\%, which is better than all four standalone VLMs and even slightly better than the AdaFace baseline (6.05\%). At 0.01\% FMR, the FNMR is 22.02\%, which is almost identical to Gemma's standalone 21.38\%, despite the fusion having substantially better EER. Gemma appears repeatedly as one of the most useful source models. The configurations involving Qwen + Gemma as sources produce the strongest results for both Intern and Ovis. Gemma seems to provide particularly useful information to the other VLMs. This is interesting given that Gemma already has the strongest standalone AUC/EER performance.

\begin{table*}
    \centering
    \caption{Scenario A FR evaluation metrics for the 12 fusion combinations.}
    \resizebox{\textwidth}{!}{%
    \begin{tabular}{lrrrrrrrr}
    \toprule
    Model & AUC & EER & EER Threshold & d-prime & FNMR@0.1\%FMR & Threshold@0.1\%FMR & FNMR@0.01\%FMR & Threshold@0.01\%FMR \\
    \midrule
    MagFace & 0.9741 & 0.0557 & 0.1343 & 4.8431 & 0.0612 & 0.2652 & 0.0620 & 0.3201 \\
    \midrule
    Gemma & 0.9985 & 0.0130 & 0.5488 & 6.5239 & 0.0688 & 0.8745 & 0.2138 & 0.9230 \\
    \midrule
    Gemma $\leftarrow$ Intern + Ovis & 0.9962 & 0.0262 & 0.7289 & 4.8478 & 0.3224 & 0.8703 & 0.6486 & 0.9129 \\
    Gemma $\leftarrow$ Qwen + Intern & 0.9980 & 0.0173 & 0.6665 & 6.6291 & 0.1687 & 0.8738 & 0.5471 & 0.9138 \\
    Gemma $\leftarrow$ Qwen + Ovis & 0.9982 & 0.0164 & 0.7183 & 6.4134 & 0.1671 & 0.9020 & 0.5384 & 0.9527 \\
    Intern $\leftarrow$ Ovis + Gemma & 0.9988 & 0.0144 & 0.6653 & 5.7538 & 0.0893 & 0.8616 & 0.3793 & 0.9090 \\
    Intern $\leftarrow$ Qwen + Gemma & \textbf{0.9992} & \textbf{0.0117} & 0.6085 & \textbf{7.9473} & \textbf{0.0517} & 0.8618 & \textbf{0.2202} & 0.9301 \\
    Intern $\leftarrow$ Qwen + Ovis & 0.9982 & 0.0165 & 0.7330 & 6.3794 & 0.1734 & 0.9024 & 0.5643 & 0.9536 \\
    Ovis $\leftarrow$ Intern + Gemma & 0.9983 & 0.0133 & 0.6219 & 5.0433 & 0.1010 & 0.8303 & 0.2823 & 0.8860 \\
    Ovis $\leftarrow$ Qwen + Gemma & 0.9977 & 0.0122 & 0.6074 & 5.1844 & 0.0523 & 0.8618 & 0.2204 & 0.9301 \\
    Ovis $\leftarrow$ Qwen + Intern & 0.9965 & 0.0173 & 0.6665 & 4.7240 & 0.1687 & 0.8738 & 0.5471 & 0.9138 \\
    Qwen $\leftarrow$ Intern + Gemma & 0.9988 & 0.0133 & 0.6219 & 5.9138 & 0.1010 & 0.8303 & 0.2823 & 0.8860 \\
    Qwen $\leftarrow$ Intern + Ovis & 0.9962 & 0.0262 & 0.7289 & 4.8478 & 0.3224 & 0.8703 & 0.6486 & 0.9129 \\
    Qwen $\leftarrow$ Ovis + Gemma & 0.9988 & 0.0144 & 0.6653 & 5.7530 & 0.0893 & 0.8616 & 0.3793 & 0.9090 \\
    \bottomrule
    \end{tabular}
    }
    \label{tab:scenarioa}
\end{table*}

There are some exactly identical results for Gemma $\leftarrow$ Intern + Ovis and Qwen $\leftarrow$ Intern + Ovis, the reason behind this could be related to the modes receiving only scores and performing a direct average between the two scores, for this reason identical source models tend to give identical results even if the decider model is different.

\begin{figure}
    \centering
    \subfigure[]{\includegraphics[width=0.24\textwidth, alt=Gemma Scenario A DET curves.]{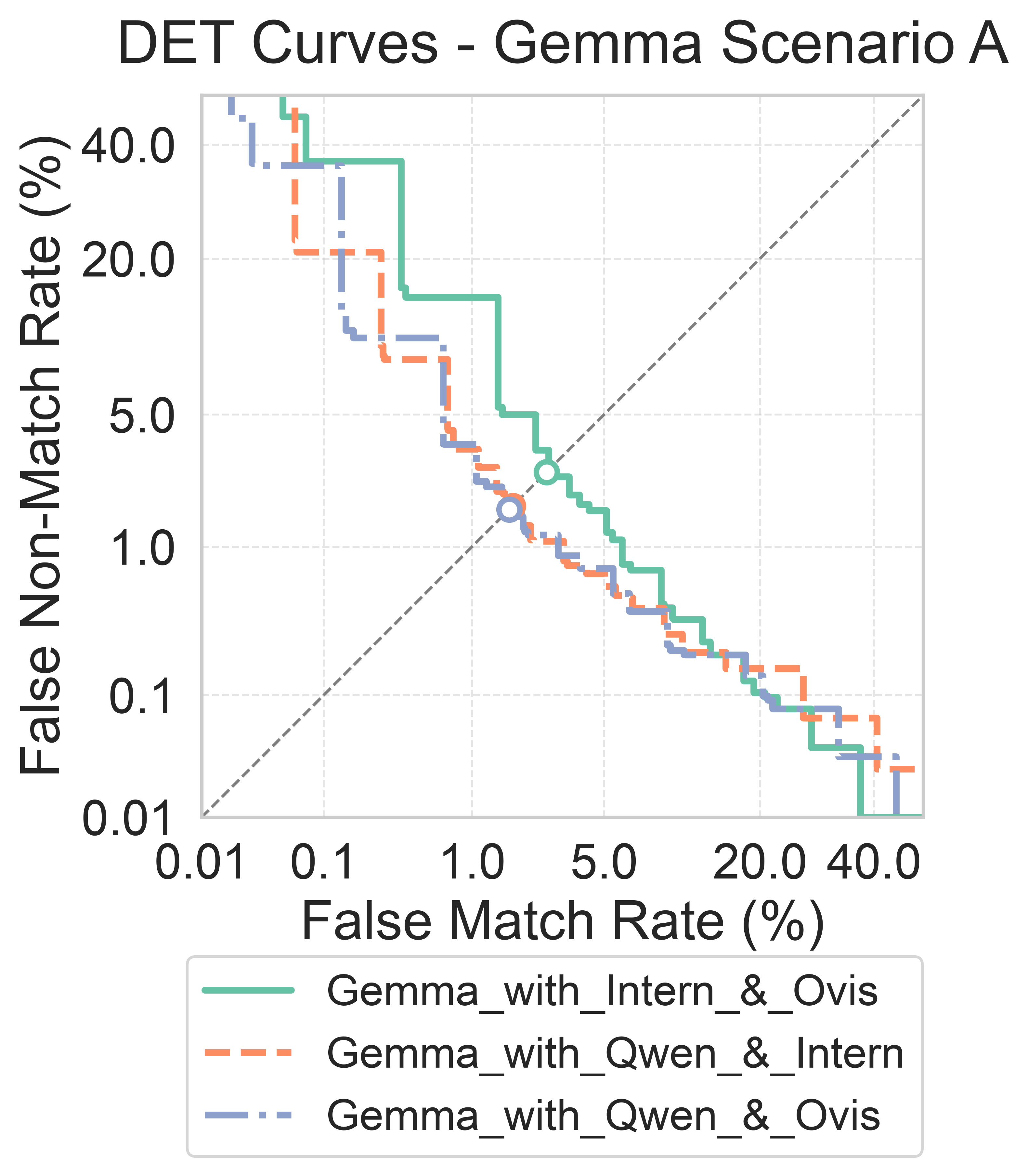}} 
    \subfigure[]{\includegraphics[width=0.24\textwidth, alt=Intern Scenario A DET curves.]{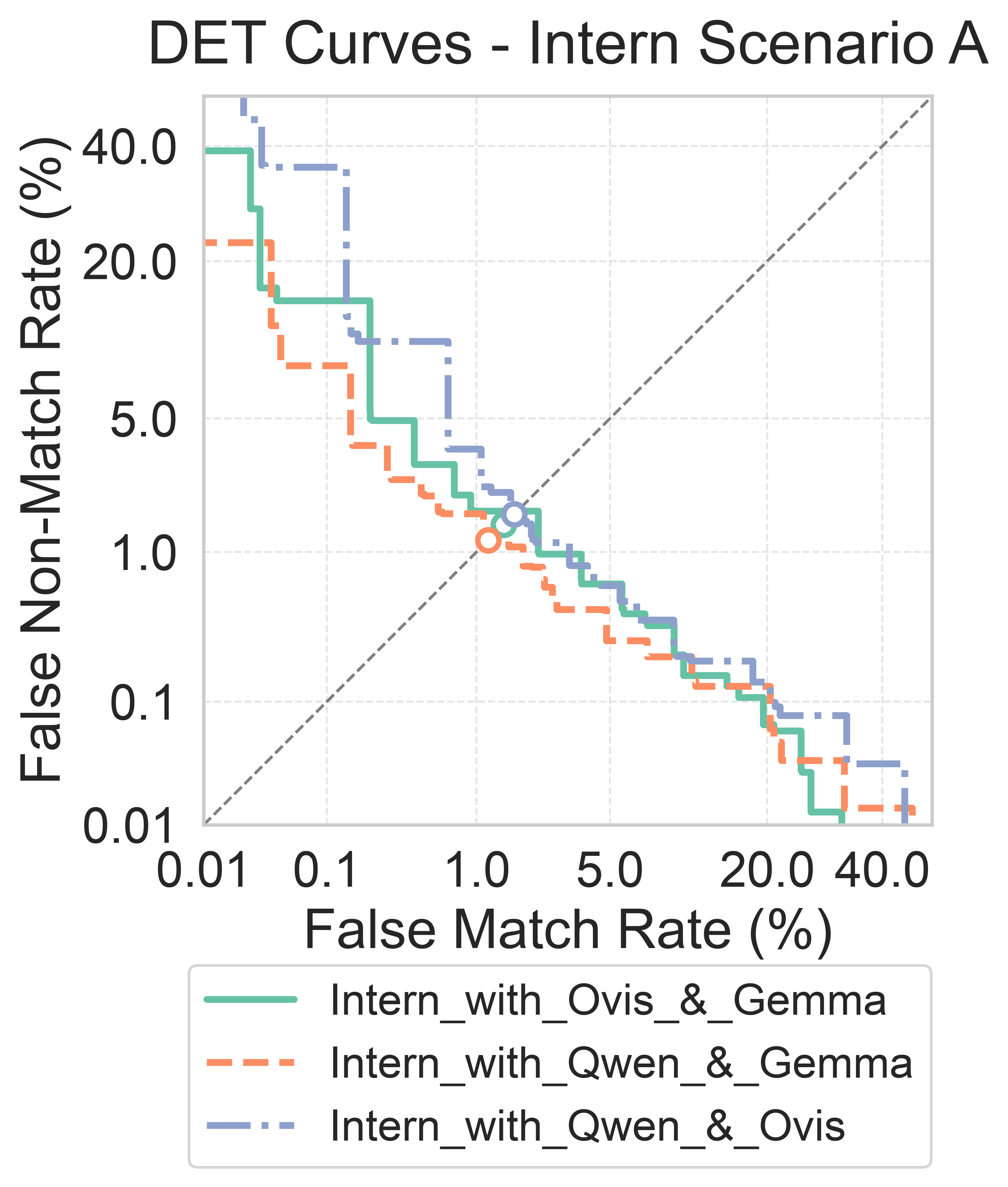}} 
    \subfigure[]{\includegraphics[width=0.24\textwidth, alt=Ovis Scenario A DET curves.]{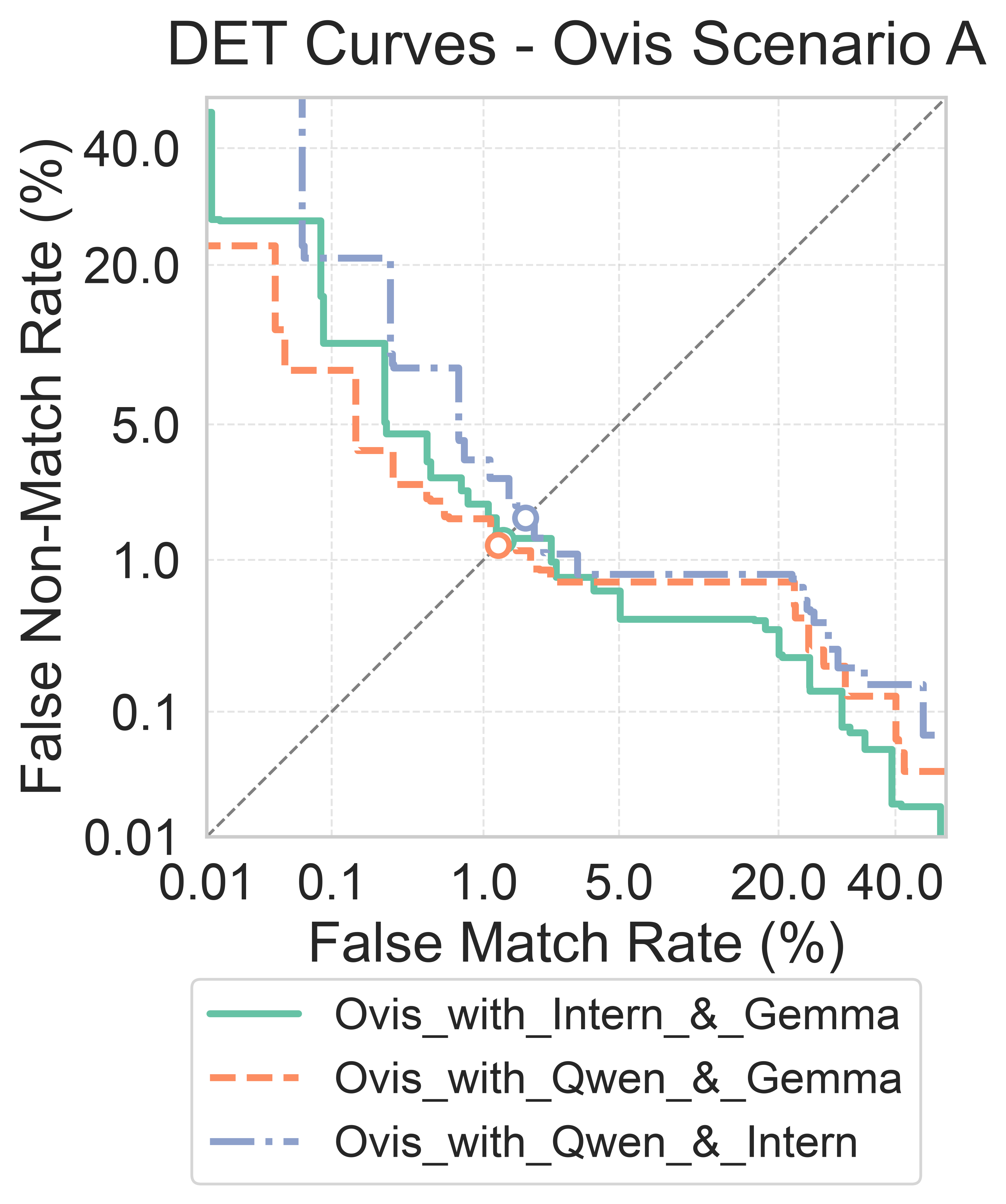}}
    \subfigure[]{\includegraphics[width=0.24\textwidth, alt=Qwen Scenario A DET curves.]{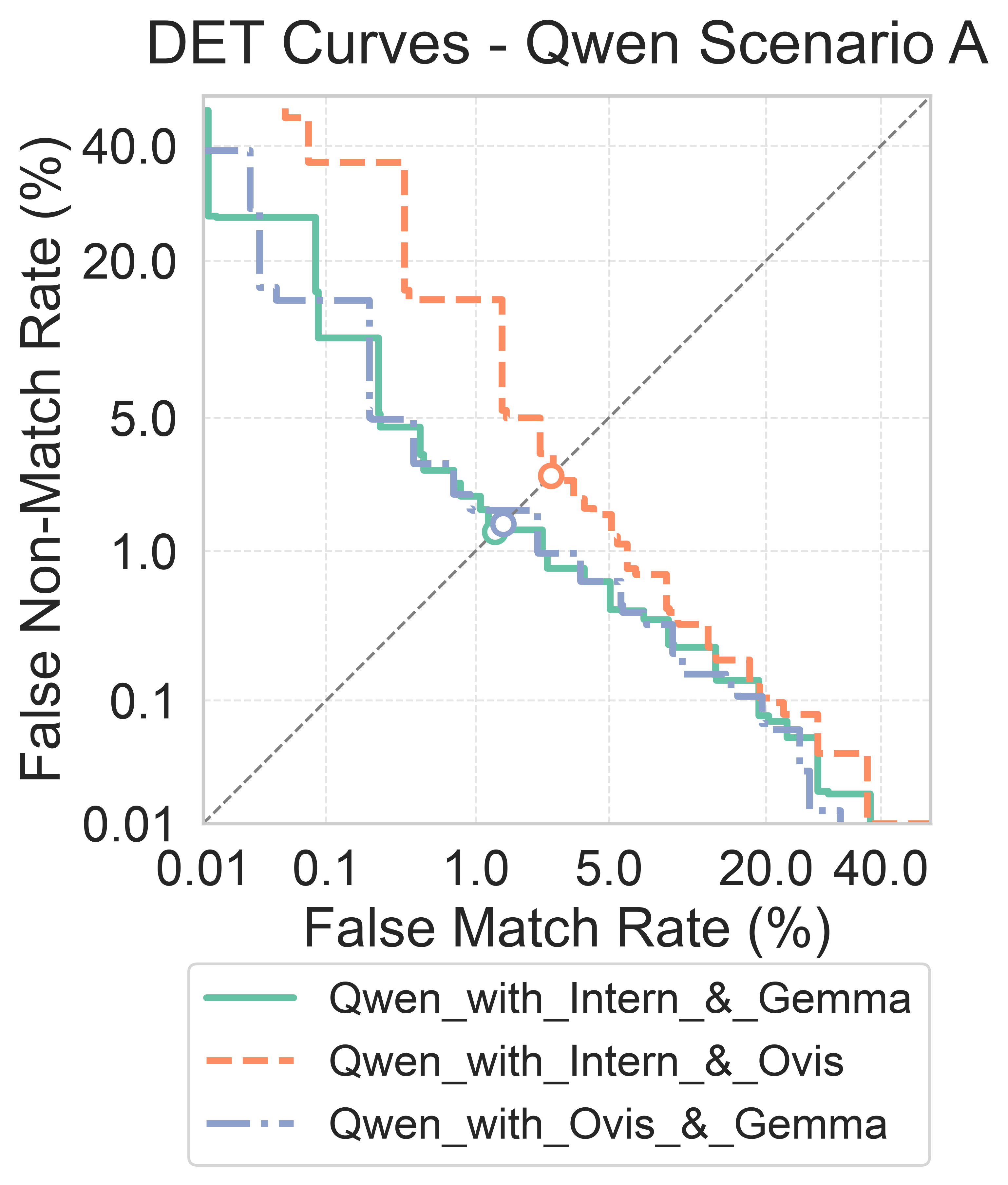}}
    \caption{Scenario A DET curves presented are grouped by decider model: (a) Gemma (b) Intern (c) Ovis and (d) Qwen.}
    \label{fig:scenarioa}
\end{figure}

In Figure \ref{fig:scenarioa} the DET curves for the 12 fusion configurations generally exhibit smoother behaviour than those of the standalone models, with fewer pronounced step-like transitions across the evaluated operating range. This suggests that combining the decisions of multiple VLMs provides a finer-grained distribution of verification scores, allowing for more gradual changes in the false match and false non-match rates as the decision threshold varies. A further observation is that configurations using Qwen and Gemma as the source models tend to produce lower DET curves around the EER operating point, consistent with the superior EER values observed for several of these configurations. In particular, the combinations involving Qwen and Gemma provide a favorable separation between genuine and impostor pairs, resulting in lower error rates around the equal-error operating point. However, the differences between curves are generally more pronounced around the EER region than at the most stringent FMR operating points, indicating that the improvement in overall discrimination does not necessarily translate into an equivalent improvement in the extreme low-FMR region.

\subsubsection{Scenario B}

In Scenario B the similarity scores and justifications of two source models are provided to a decider model. In Table \ref{tab:scenariob}, it is shown that the best configuration changes slightly compared to Scenario A, before Intern $\leftarrow$ Qwen + Gemma achieve the lowest EER of 1.17\%, whereas in Scenario B, Ovis $\leftarrow$ Qwen + Gemma achieve the lowest EER at 1.22\%. However, these differences are very small. Scenario B produces results extremely close to Scenario A. For most configurations, the changes in AUC, EER, and FNMR are small. This suggests that, at least under the current fusion methodology, the numerical scores already contain most of the information used by the decider model to make the final decision. This observation should not be interpreted as indicating that the justifications are without value. Their potential contribution lies primarily in improving the interpretability of the resulting decision, rather than necessarily improving verification accuracy. This distinction is particularly relevant to the objective of explainable FR, where explanatory information may be valuable even when it does not directly translate into improved FR performance, as will be analysed hereafter.

\begin{table*}
    \centering
    \caption{Scenario B FR evaluation metrics for the 12 fusion combinations.}
    \resizebox{\textwidth}{!}{%
    \begin{tabular}{lrrrrrrrr}
    \toprule
    Model & AUC & EER & EER Threshold & d-prime & FNMR@0.1\%FMR & Threshold@0.1\%FMR & FNMR@0.01\%FMR & Threshold@0.01\%FMR \\
    \midrule
    MagFace & 0.9741 & 0.0557 & 0.1343 & 4.8431 & 0.0612 & 0.2652 & 0.0620 & 0.3201 \\
    \midrule
    Gemma & 0.9985 & 0.0130 & 0.5488 & 6.5239 & 0.0688 & 0.8745 & 0.2138 & 0.9230 \\
    \midrule
    Gemma $\leftarrow$ Intern + Ovis & 0.9962 & 0.0261 & 0.7260 & 4.8457 & 0.3239 & 0.8705 & 0.6485 & 0.9129 \\
    Gemma $\leftarrow$ Qwen + Intern & 0.9977 & 0.0173 & 0.6676 & 6.6205 & 0.1652 & 0.8709 & 0.5458 & 0.9138 \\
    Gemma $\leftarrow$ Qwen + Ovis & 0.9978 & 0.0164 & 0.7194 & 6.4181 & 0.1670 & 0.9020 & 0.5383 & 0.9527 \\
    Intern $\leftarrow$ Ovis + Gemma & 0.9988 & 0.0143 & 0.6806 & 5.6971 & 0.0900 & 0.8620 & 0.3784 & 0.9090 \\
    Intern $\leftarrow$ Qwen + Gemma & 0.9991 & 0.0124 & 0.6170 & 7.6767 & 0.0523 & 0.8587 & 0.2186 & 0.9344 \\
    Intern $\leftarrow$ Qwen + Ovis & 0.9982 & 0.0164 & 0.7187 & 6.5003 & 0.1670 & 0.9020 & 0.6949 & 0.9568 \\
    Ovis $\leftarrow$ Intern + Gemma & 0.9988 & 0.0137 & 0.6252 & 5.8825 & 0.1029 & 0.8350 & 0.2817 & 0.8804 \\
    Ovis $\leftarrow$ Qwen + Gemma & \textbf{0.9992} & \textbf{0.0122} & 0.6074 & \textbf{7.9377} & \textbf{0.0523} & 0.8535 & \textbf{0.2201} & 0.9258 \\
    Ovis $\leftarrow$ Qwen + Intern & 0.9980 & 0.0173 & 0.6632 & 6.6249 & 0.1654 & 0.8569 & 0.5468 & 0.9166 \\
    Qwen $\leftarrow$ Intern + Gemma & 0.9987 & 0.0133 & 0.6215 & 5.8228 & 0.0977 & 0.8091 & 0.2784 & 0.8853 \\
    Qwen $\leftarrow$ Intern + Ovis & 0.9962 & 0.0262 & 0.7260 & 4.7805 & 0.3238 & 0.8664 & 0.6690 & 0.9259 \\
    Qwen $\leftarrow$ Ovis + Gemma & 0.9988 & 0.0145 & 0.6621 & 5.6830 & 0.0890 & 0.8616 & 0.3278 & 0.9090 \\
    \bottomrule
    \end{tabular}
    }
    \label{tab:scenariob}
\end{table*}

\begin{figure}
    \centering
    \subfigure[]{\includegraphics[width=0.24\textwidth, alt=DET curves for Gemma scenario B.]{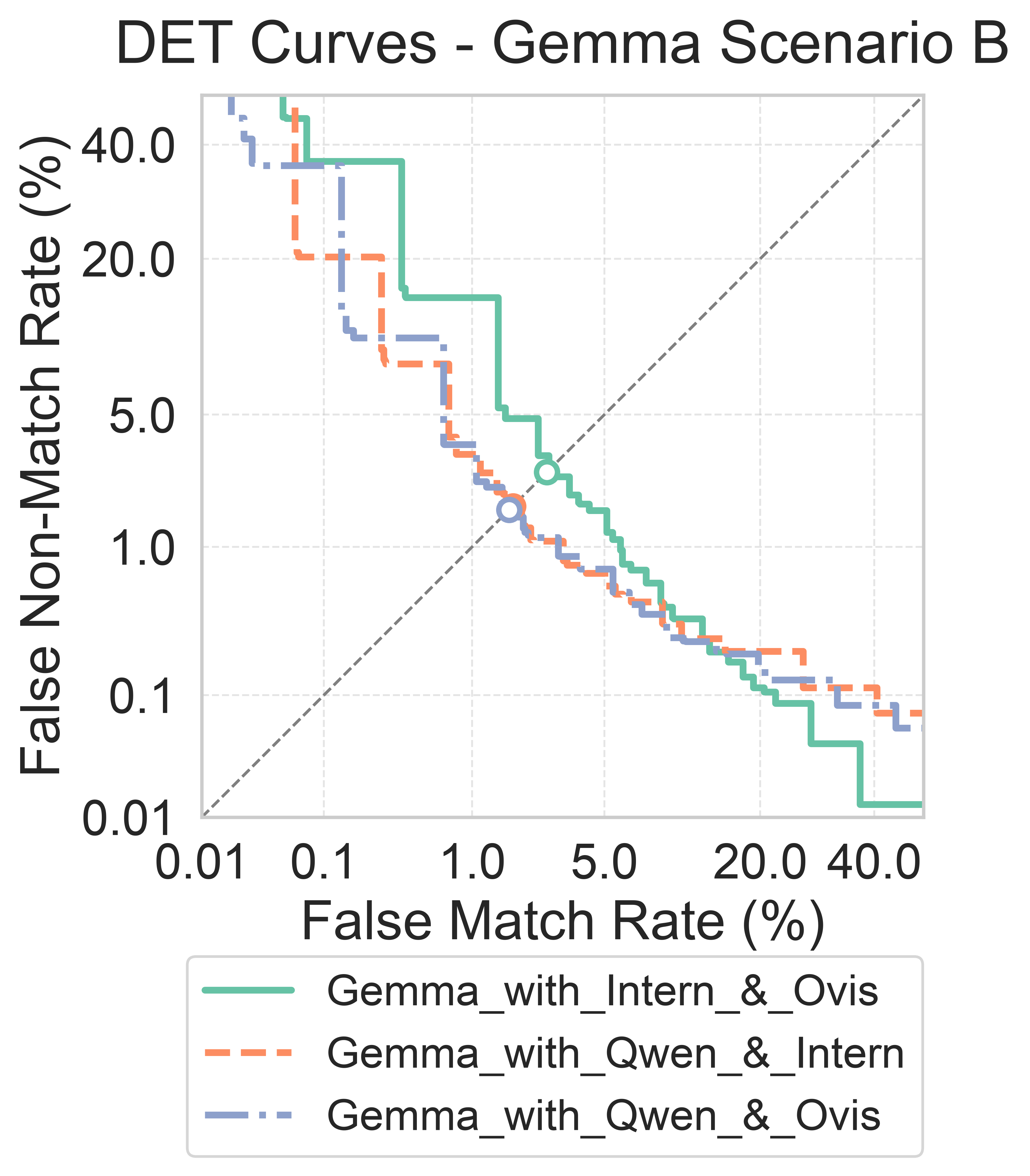}} 
    \subfigure[]{\includegraphics[width=0.24\textwidth, alt=DET curves for Intern scenario B.]{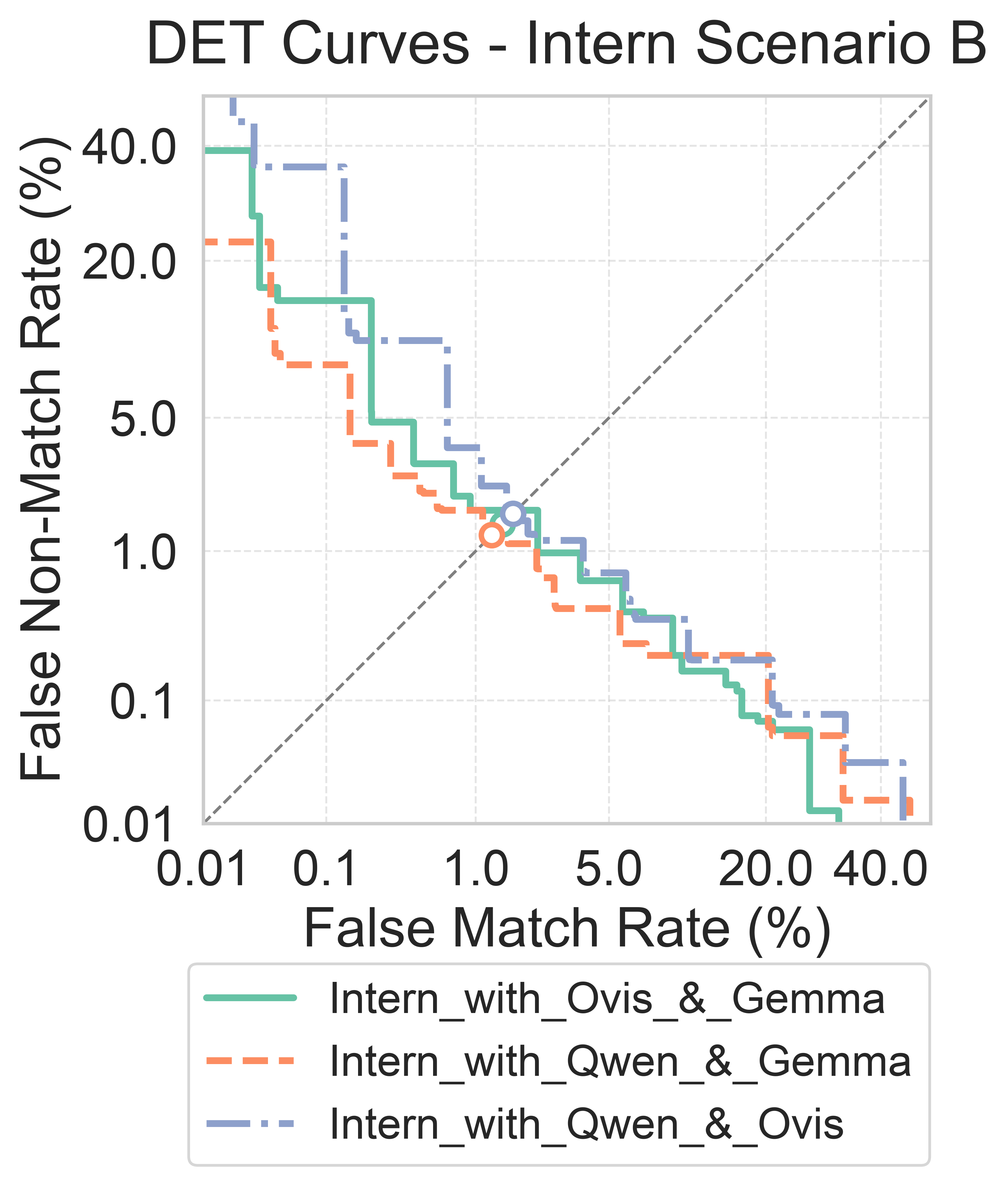}} 
    \subfigure[]{\includegraphics[width=0.24\textwidth, alt=DET curves for Ovis scenario B.]{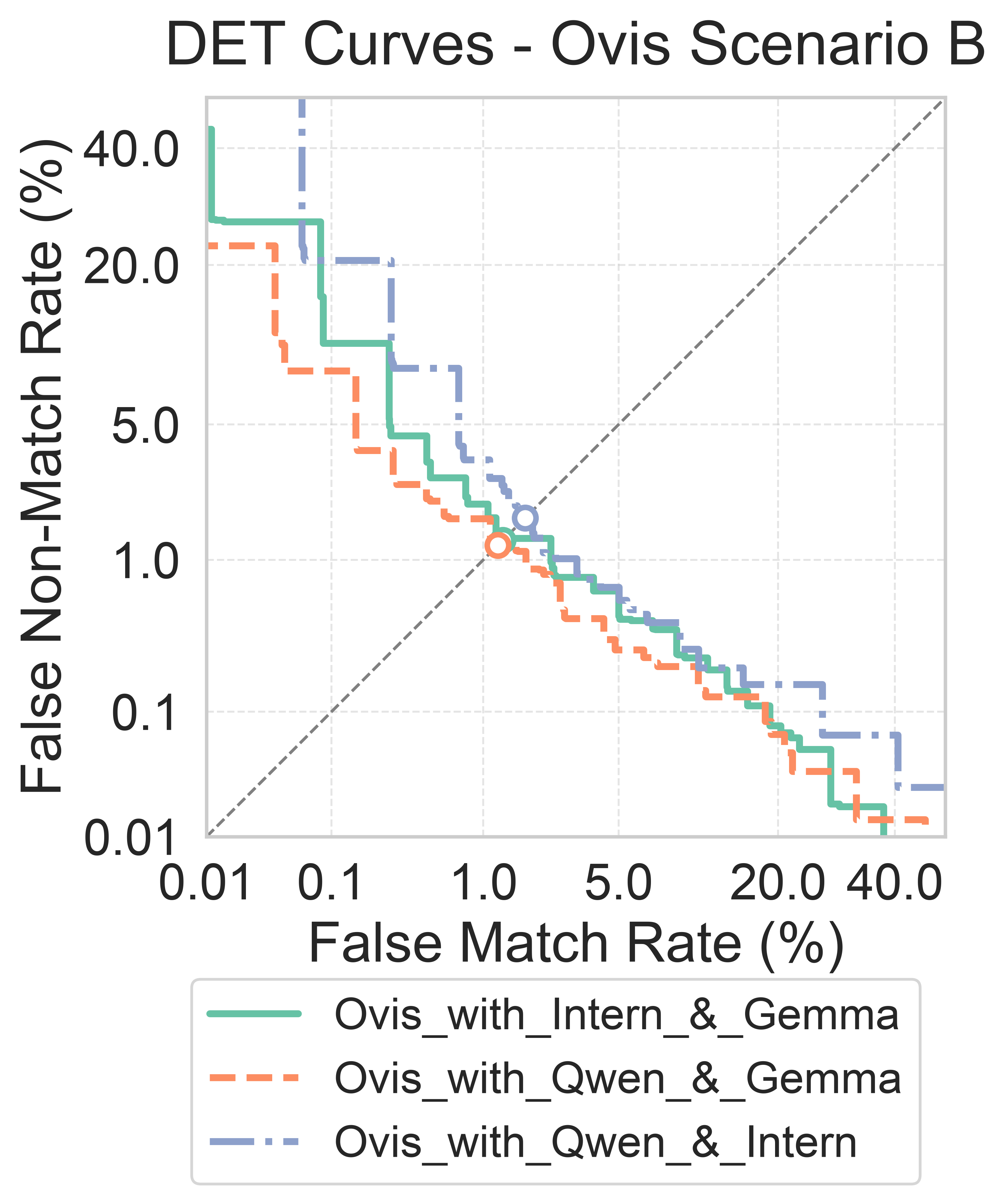}}
    \subfigure[]{\includegraphics[width=0.24\textwidth, alt=DET curves for Qwen scenario B.]{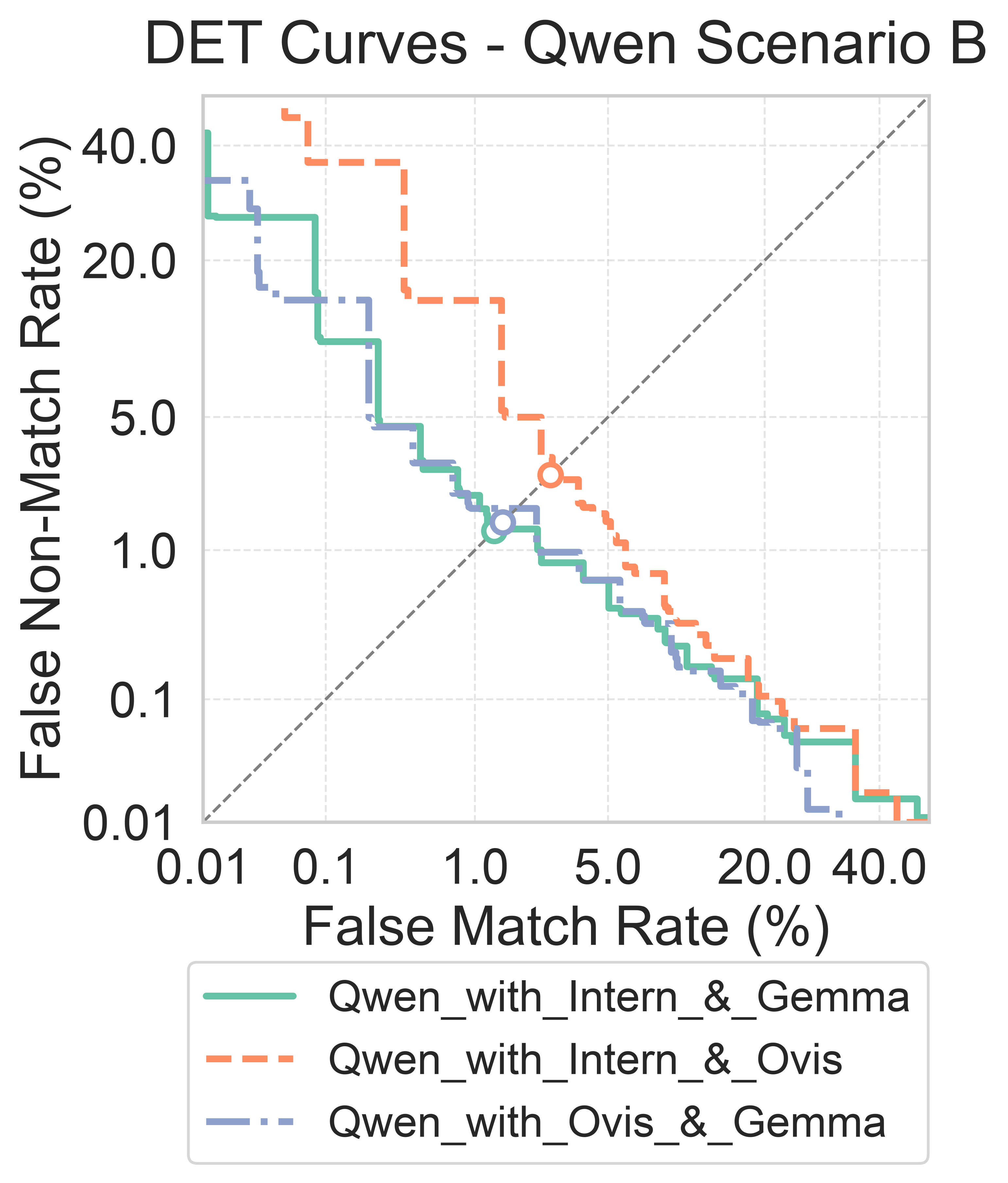}}
    \caption{Scenario B DET curves presented are grouped by decider model: (a) Gemma (b) Intern (c) Ovis (d) Qwen.}
    \label{fig:scenariob}
\end{figure}

In Figure \ref{fig:scenariob}, the DET curves obtained in Scenario B are highly similar to those observed in Scenario A, with no substantial change in their overall shape or relative ordering. This further supports the observation that incorporating the source models' justifications in addition to their scores does not significantly alter the verification behaviour. In particular, the curves remain generally smoother than those of the standalone models, while configurations using Qwen and Gemma as source models continue to exhibit favorable performance around the EER operating point.

\subsubsection{Scenario C}

In scenario C, the decider model receives the image pair and similarity scores from two source models. As shown in Table \ref{tab:scenarioc}, the results suggest that visual information can provide substantial additional information, but the benefit is highly dependent on which model performs the final fusion. The strongest results in Scenario C are obtained when Gemma is the decider model. The best EER, 1.08\%, (Gemma $\leftarrow$ Qwen + Intern) is an improvement over Gemma's standalone EER of 1.30\%, and also over the corresponding Scenario A/B configurations. More importantly, the low-FMR performance improves compared with the standalone Gemma model. The improvement is clear at 0.1\% FMR, but not at 0.01\% FMR. In fact, the strictest operating point becomes worse. One of the most interesting results is that Intern as the decider model performs substantially worse when images are added. For Intern, providing the images appears to interfere with the score-based decision fusion. The same general behaviour appears for Ovis, which performs considerably worse in several image-based configurations. Qwen's results remain relatively strong, they are broadly comparable to its Scenario A/B results, although they don't surpass the best score-only configurations. Gemma seems to benefit most from access to the images.

\begin{table*}
    \centering
    \caption{Scenario C FR evaluation metrics for the 12 fusion combinations.}
    \resizebox{\textwidth}{!}{%
    \begin{tabular}{lrrrrrrrr}
    \toprule
    Model & AUC & EER & EER Threshold & d-prime & FNMR@0.1\%FMR & Threshold@0.1\%FMR & FNMR@0.01\%FMR & Threshold@0.01\%FMR \\
    \midrule
    MagFace & 0.9741 & 0.0557 & 0.1343 & 4.8431 & 0.0612 & 0.2652 & 0.0620 & 0.3201 \\
    \midrule
    Gemma & 0.9985 & 0.0130 & 0.5488 & 6.5239 & 0.0688 & 0.8745 & 0.2138 & 0.9230 \\
    \midrule
    Gemma $\leftarrow$ Intern + Ovis & 0.9987 & 0.0129 & 0.6759 & 6.6103 & 0.0878 & 0.8782 & 0.2906 & 0.9333 \\
    Gemma $\leftarrow$ Qwen + Intern & 0.9988 & \textbf{0.0108} & 0.6627 & 7.5194 & 0.0649 & 0.8753 & \textbf{0.2759} & 0.9330 \\
    Gemma $\leftarrow$ Qwen + Ovis & \textbf{0.9989} & 0.0112 & 0.6685 & 7.4923 & \textbf{0.0626} & 0.8941 & 0.3393 & 0.9362 \\
    Intern $\leftarrow$ Ovis + Gemma & 0.9969 & 0.0239 & 0.7308 & 5.5130 & 0.2319 & 0.9028 & 0.5054 & 0.9451 \\
    Intern $\leftarrow$ Qwen + Gemma & 0.9979 & 0.0171 & 0.6998 & 6.3021 & 0.1584 & 0.9003 & 0.3807 & 0.9184 \\
    Intern $\leftarrow$ Qwen + Ovis & 0.9973 & 0.0193 & 0.7145 & 6.0313 & 0.2420 & 0.9076 & 0.5280 & 0.9532 \\
    Ovis $\leftarrow$ Intern + Gemma & 0.9958 & 0.0283 & 0.6555 & 4.9163 & 0.2651 & 0.8388 & 0.7479 & 0.9126 \\
    Ovis $\leftarrow$ Qwen + Gemma & 0.9980 & 0.0153 & 0.6465 & 6.0796 & 0.1294 & 0.8516 & 0.4385 & 0.9236 \\
    Ovis $\leftarrow$ Qwen + Intern & 0.9965 & 0.0210 & 0.6503 & 5.7462 & 0.3200 & 0.9006 & 0.6349 & 0.9126 \\
    Qwen $\leftarrow$ Intern + Gemma & 0.9981 & 0.0133 & 0.7014 & \textbf{7.9049} & 0.1700 & 0.9225 & 0.5331 & 0.9512 \\
    Qwen $\leftarrow$ Intern + Ovis & 0.9974 & 0.0163 & 0.7538 & 7.2747 & 0.2045 & 0.9268 & 0.6422 & 0.9481 \\
    Qwen $\leftarrow$ Ovis + Gemma & 0.9980 & 0.0138 & 0.7539 & 7.8143 & 0.1569 & 0.9267 & 0.3943 & 0.9535 \\
    \bottomrule
    \end{tabular}
    }
    \label{tab:scenarioc}
\end{table*}

\begin{figure}
    \centering
    \subfigure[]{\includegraphics[width=0.240\textwidth, alt=DET curves for Gemma scenario C.]{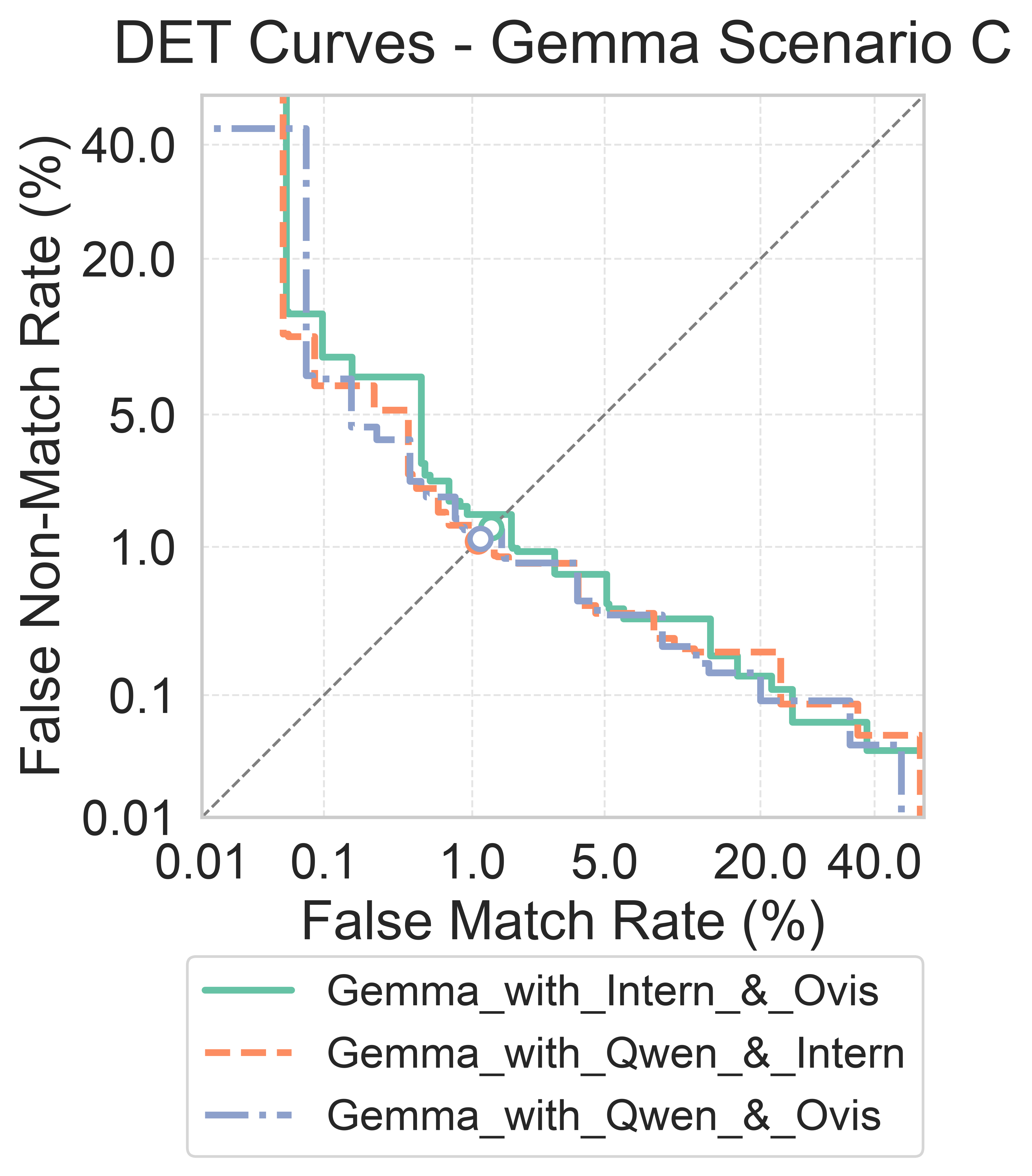}} 
    \subfigure[]{\includegraphics[width=0.240\textwidth, alt=DET curves for Intern scenario C.]{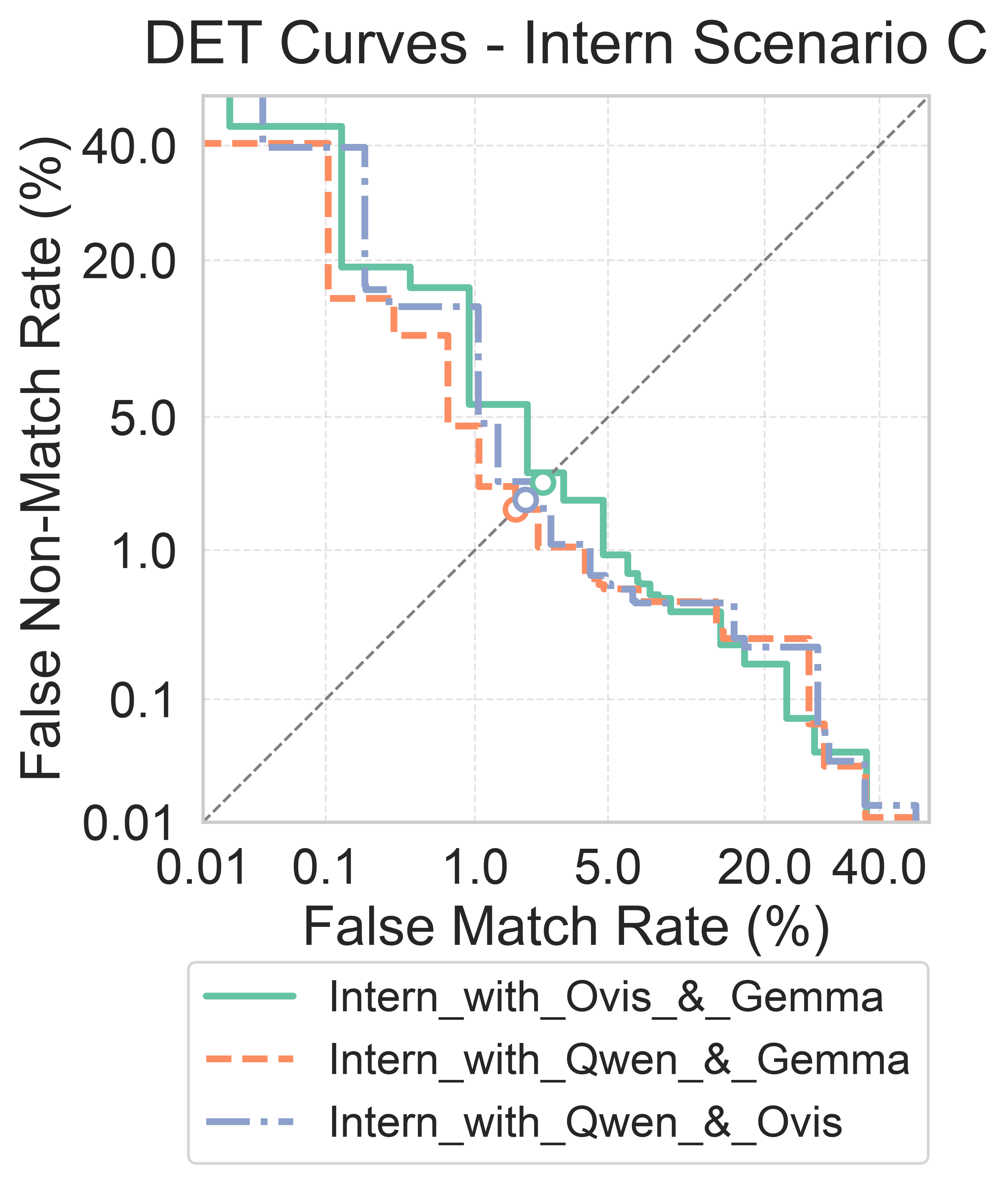}} 
    \subfigure[]{\includegraphics[width=0.240\textwidth, alt=DET curves for Ovis scenario C.]{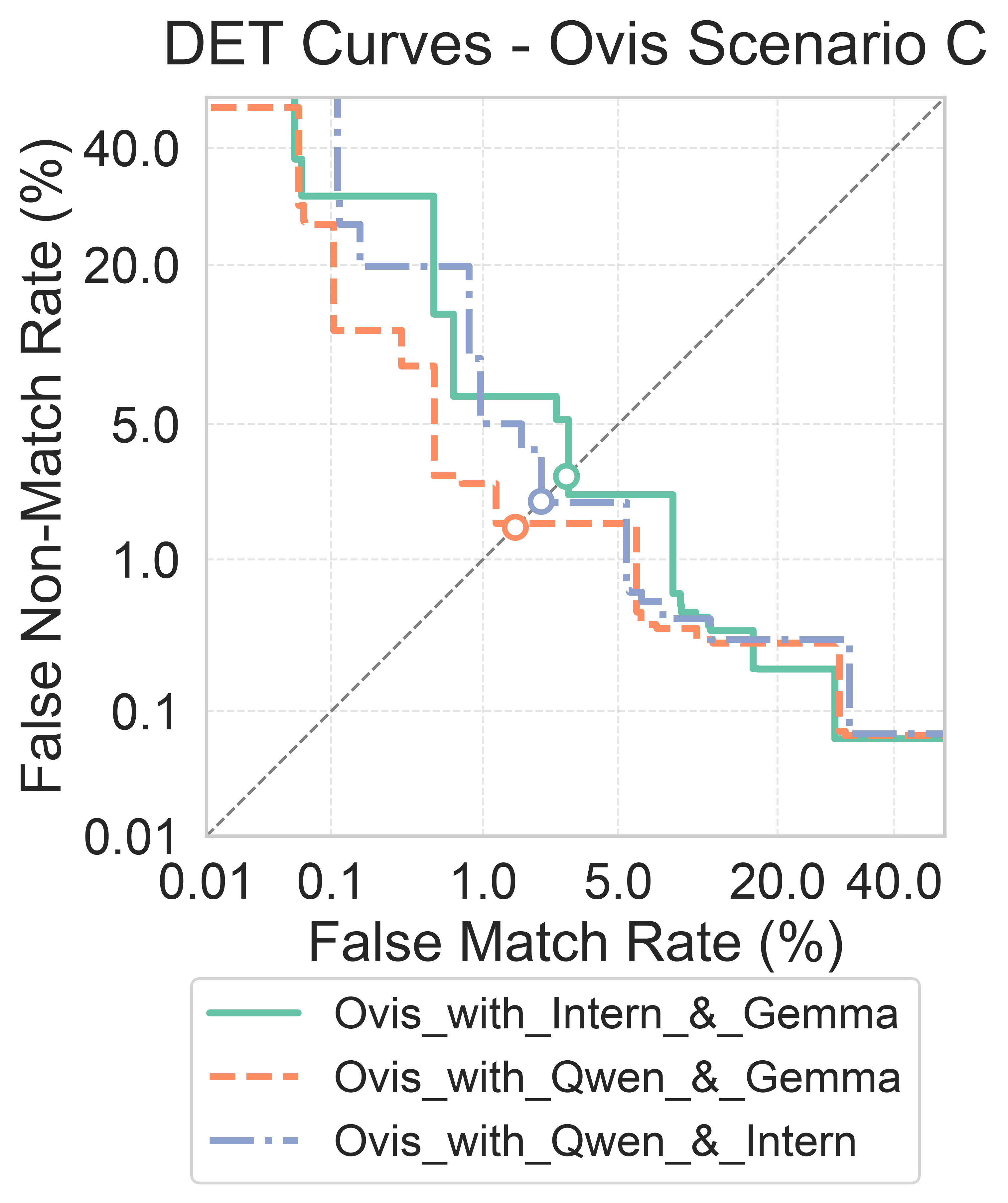}}
    \subfigure[]{\includegraphics[width=0.240\textwidth, alt=DET curves for Qwen scenario C.]{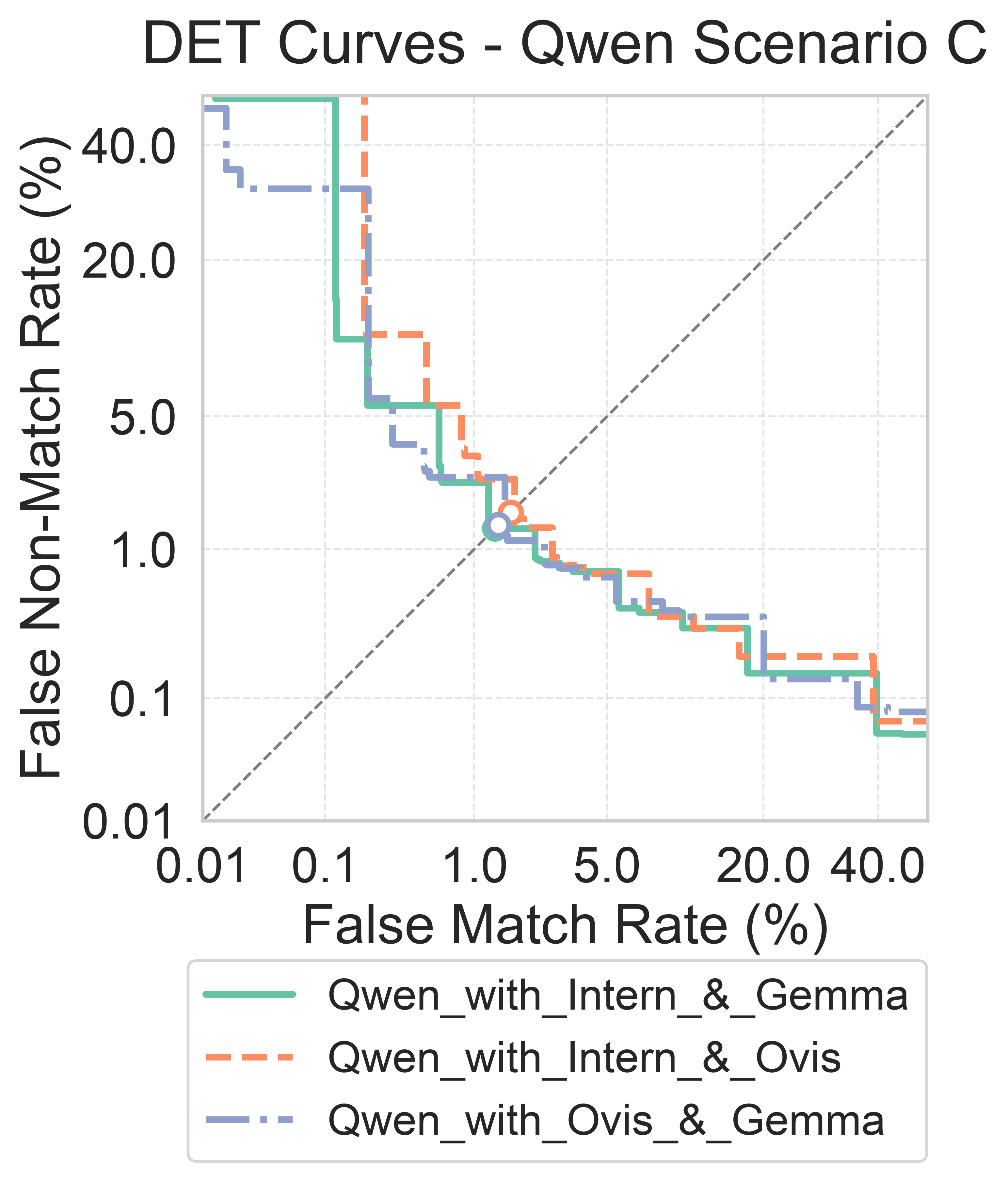}}
    \caption{Scenario C DET curves presented are grouped by decider model: (a) Gemma (b) Intern (c) Ovis (d) Qwen.}
    \label{fig:scenarioc}
\end{figure}

In Figure \ref{fig:scenarioc}, the DET curves for Scenario C show that the three configurations using Gemma as the decider model exhibit very similar behaviour, with all three achieving the lowest EERs among the evaluated configurations. Qwen follows a similar pattern, although with slightly greater variation between source combinations. For Intern and Ovis, the best-performing configurations are those that combine Gemma and Qwen as source models, consistent with the numerical EER results. Overall, the curves reinforce the observation that the choice of decider model and, particularly, the combination of source models have a substantial influence on fusion performance.

\subsubsection{Scenario D}

In scenario D the decider model has access to the face image pair, the similarity scores and justification of two source models. In Table \ref{tab:scenariod} the strongest result is Gemma $\leftarrow$ Qwen + Intern, with an AUC of 0.9987 and an EER of 1.06\%, closely followed by Gemma $\leftarrow$ Qwen + Ovis at 1.13\% EER. These results are very similar to those obtained in Scenario C, where the images and scores were provided without the justifications. This suggests that, from a verification-performance perspective, the additional textual justifications provide limited additional benefit when both the images and numerical scores are already available. A similar pattern can be observed for the other decider models. The best Ovis configuration achieves an EER of 1.22\%, while the best Qwen and Intern configurations achieve 1.30\% and 1.34\%, respectively. Thus, although Scenario D provides the richest information to the decider model, its overall performance remains strongly dependent on the decider model and source model combination.

\begin{table*}
    \centering
    \caption{Scenario D FR evaluation metrics for the 12 fusion combinations.}
    \resizebox{\textwidth}{!}{%
    \begin{tabular}{lrrrrrrrr}
    \toprule
    Model & AUC & EER & EER Threshold & d-prime & FNMR@0.1\%FMR & Threshold@0.1\%FMR & FNMR@0.01\%FMR & Threshold@0.01\%FMR \\
    \midrule
    MagFace & 0.9741 & 0.0557 & 0.1343 & 4.8431 & 0.0612 & 0.2652 & 0.0620 & 0.3201 \\
    \midrule
    Gemma & 0.9985 & 0.0130 & 0.5488 & 6.5239 & 0.0688 & 0.8745 & 0.2138 & 0.9230 \\
    \midrule
    Gemma $\leftarrow$ Intern + Ovis & 0.9985 & 0.0147 & 0.7130 & 5.8847 & 0.1047 & 0.8808 & 0.3595 & 0.9301 \\
    Gemma $\leftarrow$ Qwen + Intern & 0.9987 & \textbf{0.0106} & 0.7276 & \textbf{7.4262} & \textbf{0.0720} & 0.8871 & 0.3052 & 0.9339 \\
    Gemma $\leftarrow$ Qwen + Ovis & \textbf{0.9988} & 0.0113 & 0.7549 & 7.3283 & 0.0793 & 0.8973 & 0.3508 & 0.9282 \\
    Intern $\leftarrow$ Ovis + Gemma & 0.9979 & 0.0184 & 0.7474 & 5.2555 & 0.1977 & 0.9033 & 0.4756 & 0.9500 \\
    Intern $\leftarrow$ Qwen + Gemma & 0.9988 & 0.0134 & 0.7246 & 6.9593 & 0.1050 & 0.9040 & 0.3495 & 0.9621 \\
    Intern $\leftarrow$ Qwen + Ovis & 0.9981 & 0.0156 & 0.7426 & 6.0682 & 0.1855 & 0.9139 & 0.4877 & 0.9567 \\
    Ovis $\leftarrow$ Intern + Gemma & 0.9983 & 0.0179 & 0.6590 & 5.6366 & 0.1065 & 0.8411 & 0.5842 & 0.9079 \\
    Ovis $\leftarrow$ Qwen + Gemma & 0.9989 & 0.0122 & 0.6525 & 7.1729 & 0.1015 & 0.8724 & \textbf{0.2603} & 0.9167 \\
    Ovis $\leftarrow$ Qwen + Intern & 0.9978 & 0.0189 & 0.7030 & 6.0007 & 0.2001 & 0.8787 & 0.5563 & 0.9115 \\
    Qwen $\leftarrow$ Intern + Gemma & 0.9979 & 0.0130 & 0.6800 & 7.1221 & 0.1072 & 0.8959 & 0.6654 & 0.9487 \\
    Qwen $\leftarrow$ Intern + Ovis & 0.9971 & 0.0187 & 0.7591 & 6.2628 & 0.2207 & 0.9219 & 0.6855 & 0.9290 \\
    Qwen $\leftarrow$ Ovis + Gemma & 0.9974 & 0.0136 & 0.6786 & 7.0018 & 0.1619 & 0.9201 & 0.4176 & 0.9400 \\
    \bottomrule
    \end{tabular}
    }
    \label{tab:scenariod}
\end{table*}

An interesting comparison across the four scenarios is that Gemma remains particularly robust when acting as the decider model. Its best EER improves from 1.30\% as a standalone model to 1.08\% in Scenario C and 1.06\% in Scenario D. In contrast, adding images and justifications does not consistently benefit Intern or Ovis compared with score-only fusion. This reinforces the observation from Scenario C that additional information does not necessarily translate into better fusion performance, and that the ability of the decider model to effectively integrate the different information modalities is an important factor. For the strict operating points, Scenario D again shows a distinction between global performance and extreme low-FMR performance. The best configuration, Gemma $\leftarrow$ Qwen + Intern, achieves 7.20\% FNMR at 0.1\% FMR and 30.52\% at 0.01\% FMR. Thus, despite achieving an EER close to 1\%, the system still exhibits substantially higher FNMR when operating at the very low FMR of 0.01\%. This indicates that the improvements obtained through multimodal fusion are primarily reflected in overall discrimination and around the EER operating region, while the extreme impostor tail remains challenging.

\begin{figure}
    \centering
    \subfigure[]{\includegraphics[width=0.24\textwidth, alt=DET curves for Gemma scenario D.]{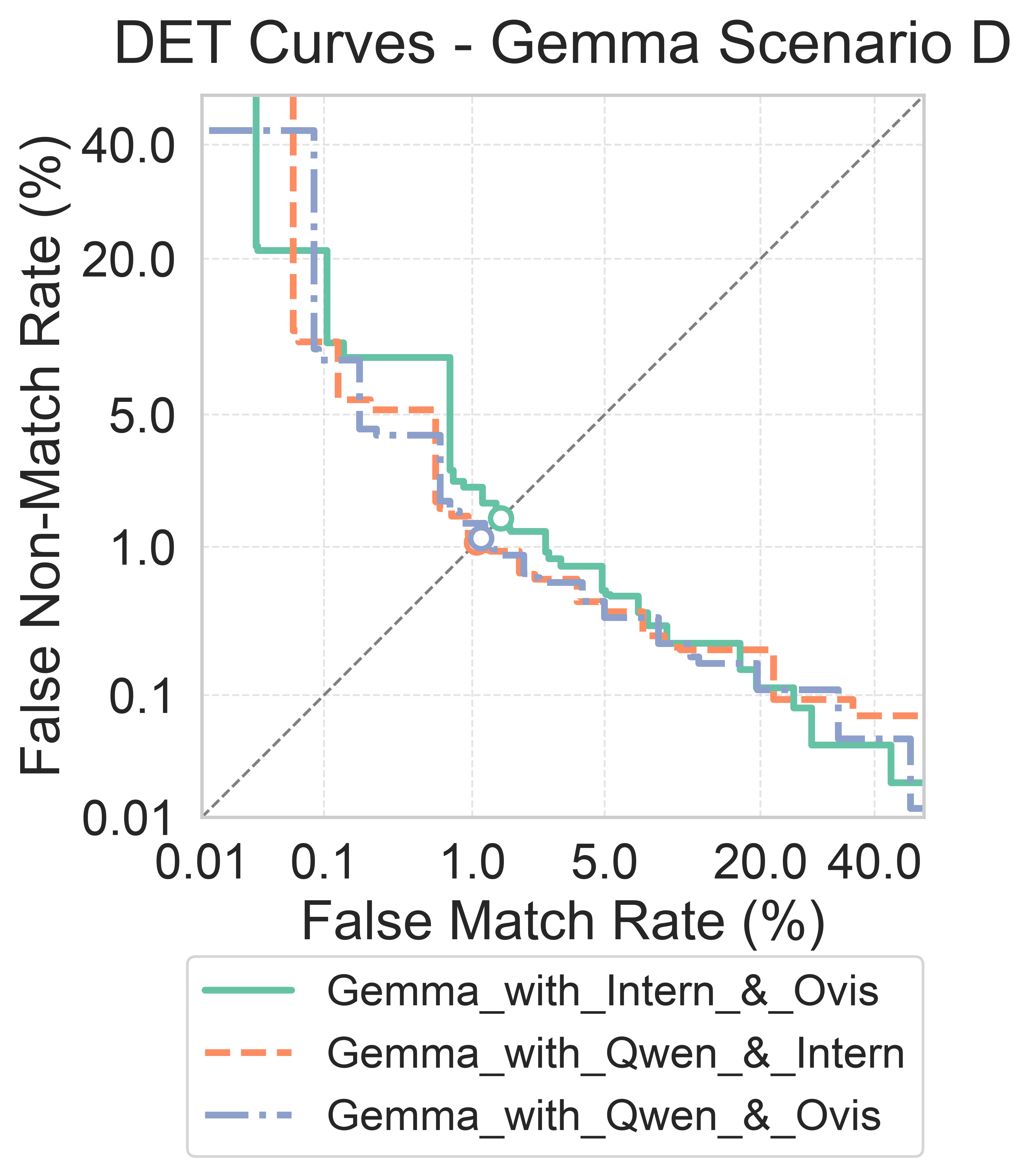}} 
    \subfigure[]{\includegraphics[width=0.24\textwidth, alt=DET curves for Intern scenario D.]{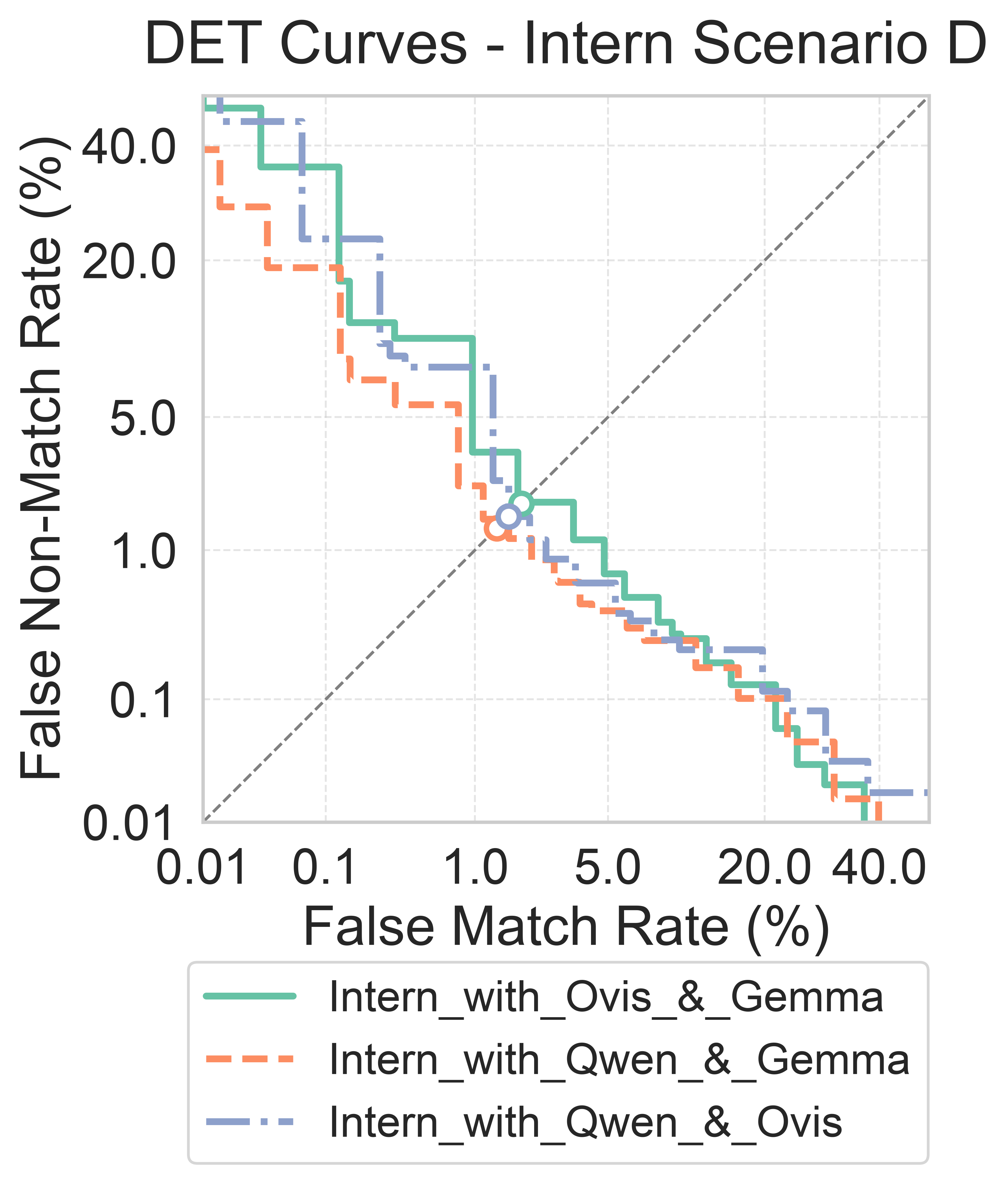}} 
    \subfigure[]{\includegraphics[width=0.24\textwidth, alt=DET curves for Ovis scenario D.]{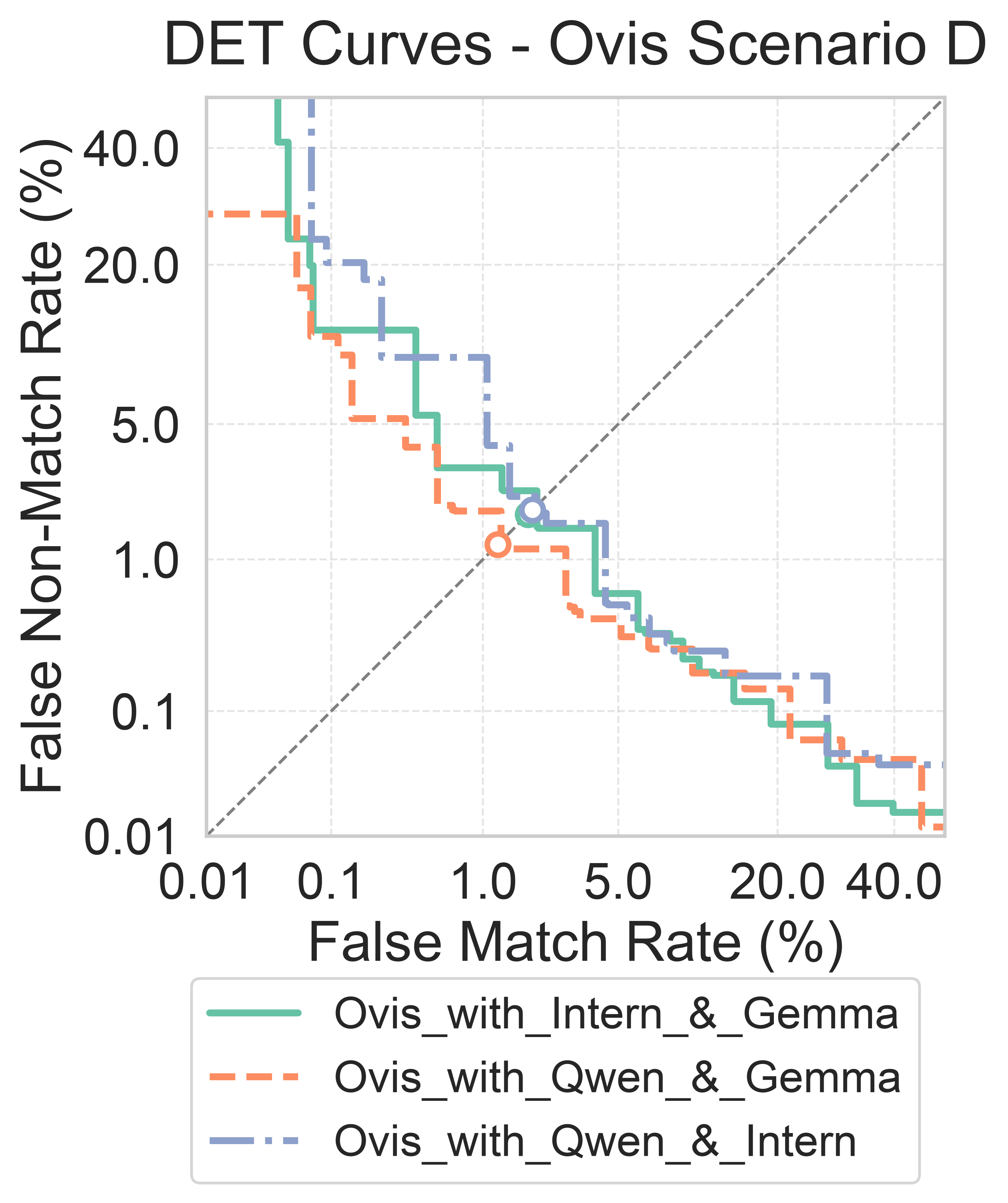}}
    \subfigure[]{\includegraphics[width=0.24\textwidth, alt=DET curves for Qwen scenario D.]{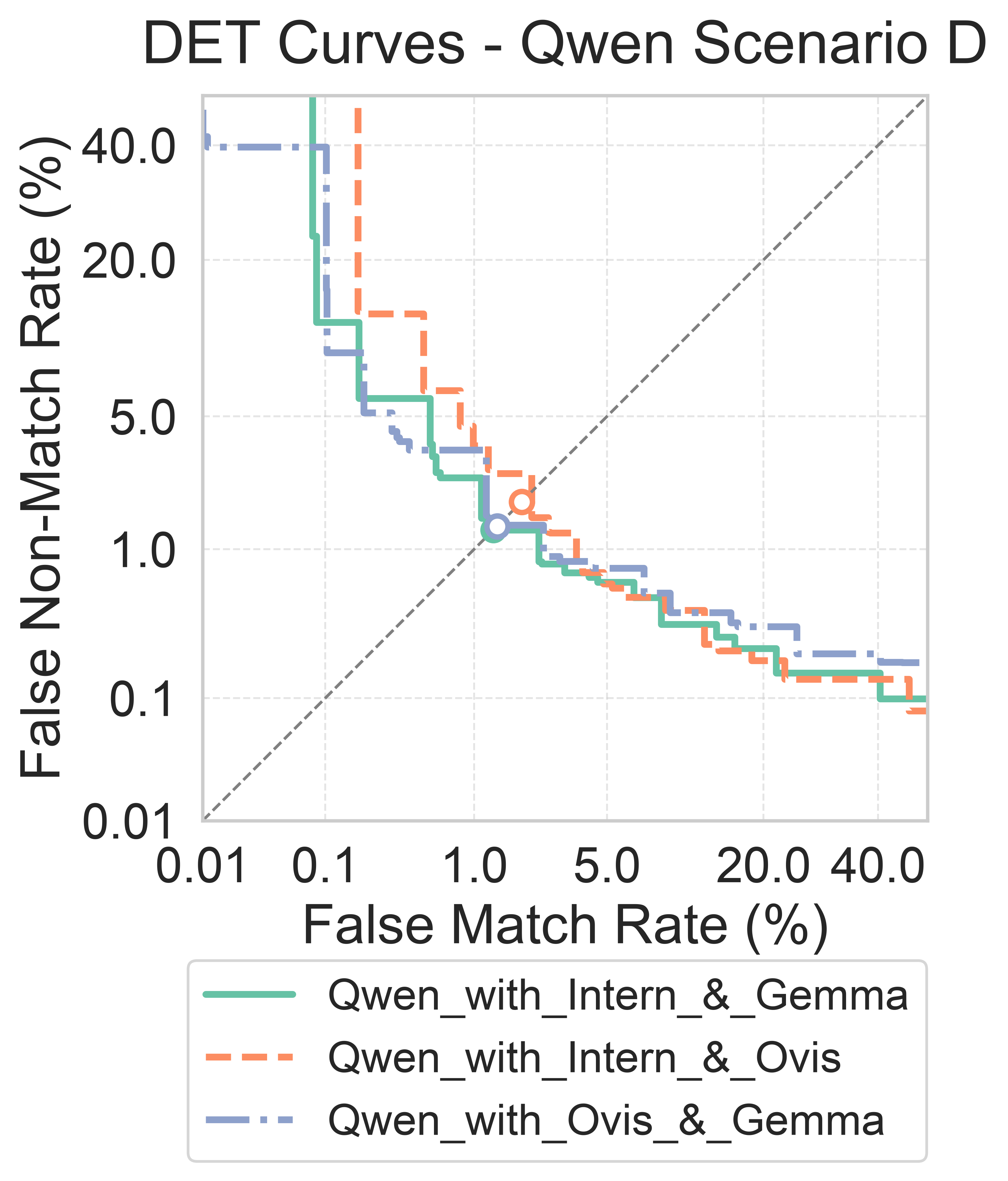}}
    \caption{Scenario D DET curves presented are grouped by decider model: (a) Gemma (b) Intern (c) Ovis (d) Qwen.}
    \label{fig:scenariod}
\end{figure}

The DET curves for Scenario D further illustrate the differences between the source model combinations. For Gemma and Qwen as decider models, a visible separation can be observed between the configurations involving Qwen and Gemma as source models and the Intern + Ovis combination, which exhibits a higher EER and correspondingly less favorable DET behaviour. For Ovis and Intern, the configuration using Qwen + Gemma as the source models consistently provides the lowest EER, reflected by the lower position of its DET curve around the equal-error operating point. These observations are consistent with the numerical results and further highlight the importance of the source model combination in determining fusion performance.

\subsection{Qualitative Analysis of Explanations}

\begin{figure*}
  \centering
    \subfigure[Gemma: \textit{The two images show a very high degree of similarity in intrinsic facial features. The eye shape, the bridge and tip of the nose, and the structure of the mouth and philtrum are nearly identical. The spatial relationship between the eyes, nose, and mouth is consistent across both images. The jawline and cheekbone structure also align closely, suggesting they are the same individual.}
]{\includegraphics[width=0.47\textwidth, alt=Image pair of two similar individuals with Gemma's justification.]{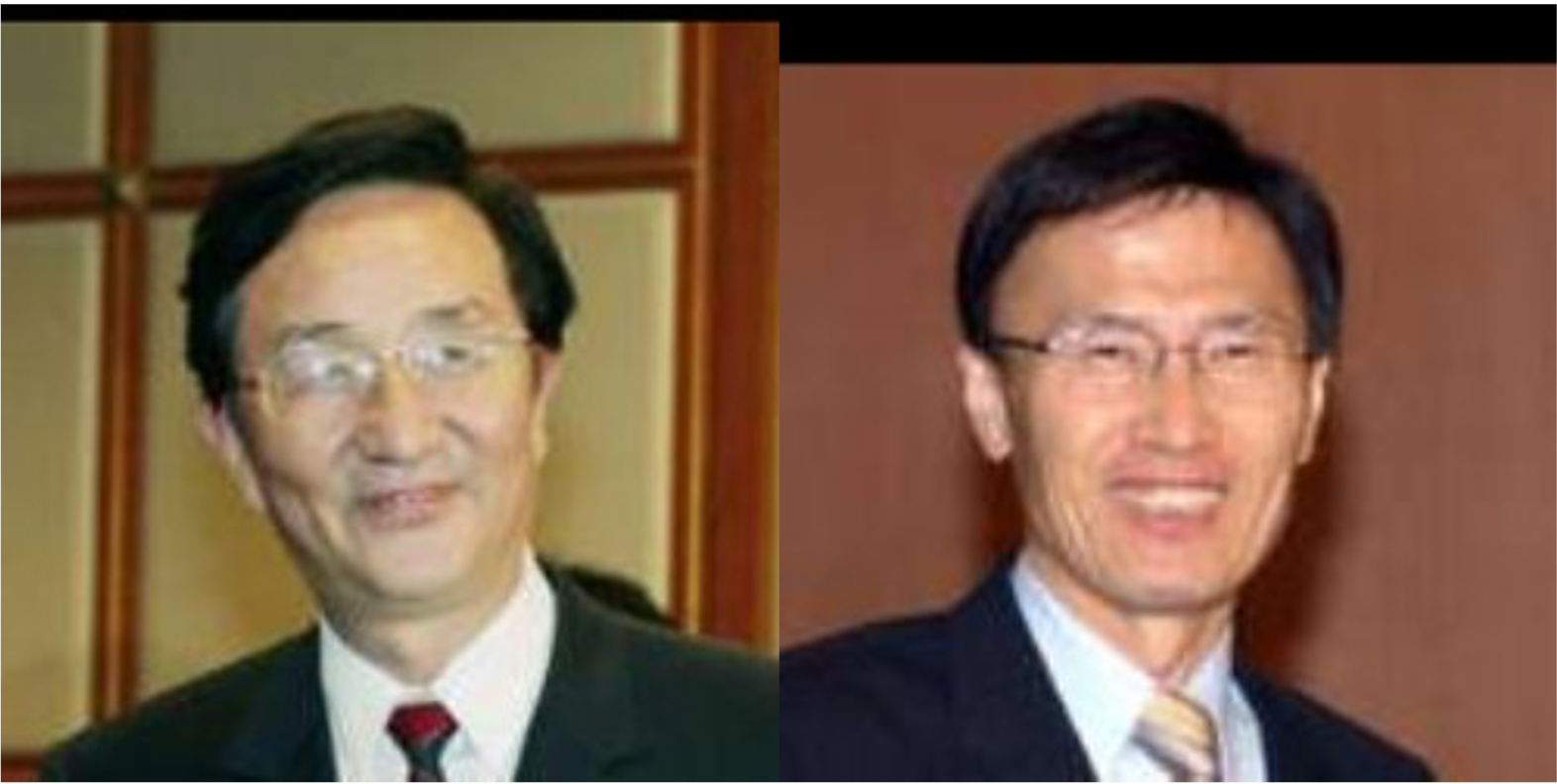}} \hfill
    \subfigure[Intern: \textit{The faces differ significantly in gender, with one being male and the other female. The facial features such as jawline, cheekbones, and overall structure are distinctively different. Additionally, the hair color and style are not similar, further indicating a lack of visual similarity.}
]{\includegraphics[width=0.47\textwidth, alt= Image pair of the same individual with Intern's justification.]{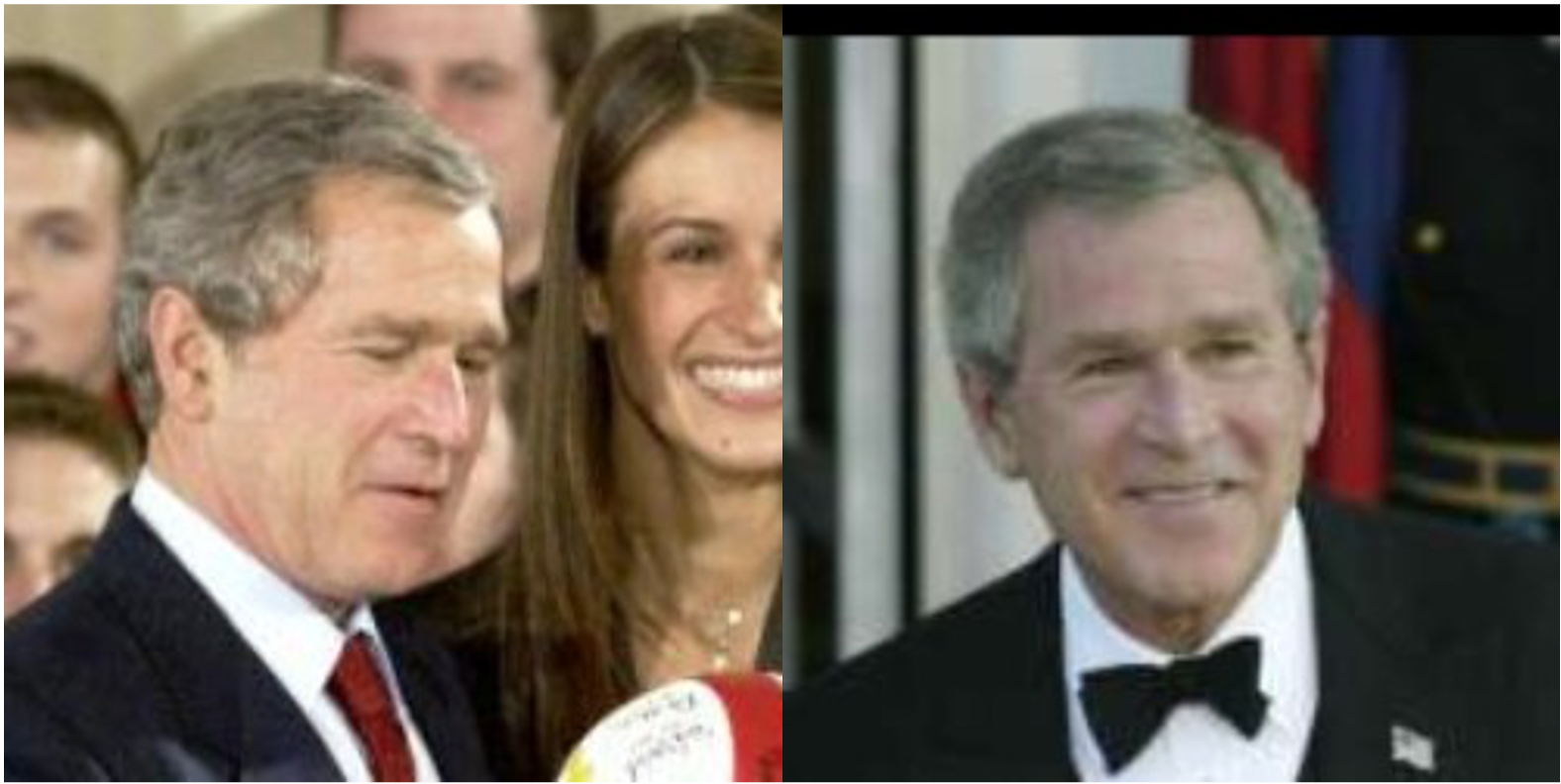}} 
    \subfigure[Ovis: \textit{The two images display a high degree of facial similarity. Both individuals have a similar structure of eyebrows, eyes, nose, and mouth. The jawline and cheekbone shapes are consistent across both images. The spatial arrangement of these features aligns closely, indicating strong visual resemblance. Minor differences in lighting and image quality do not significantly affect the overall assessment.}
]{\includegraphics[width=0.47\textwidth, alt=Image pair of two very similar individuals with Ovis' justification.]{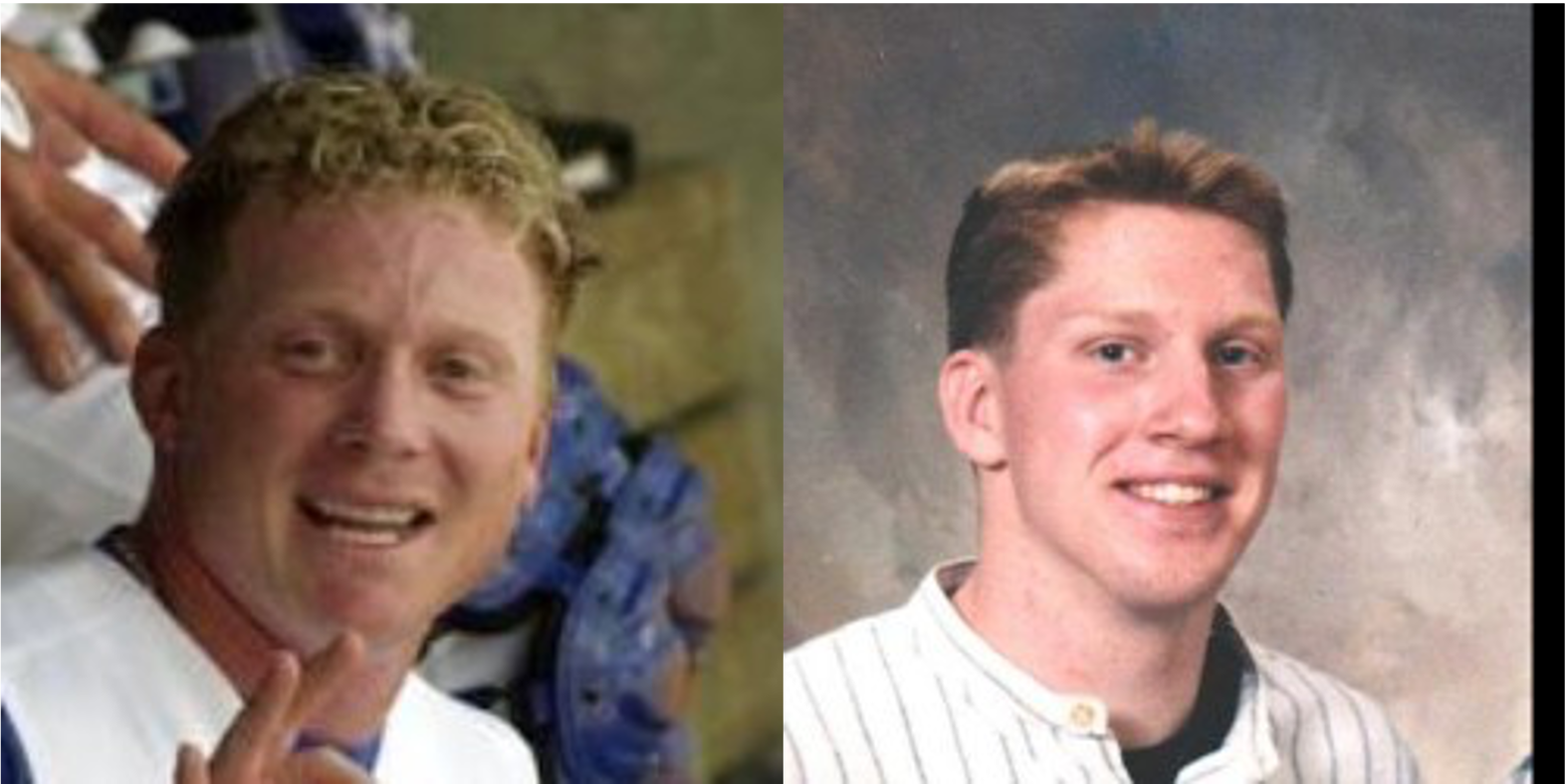}}\hfill
    \subfigure[Qwen: \textit{Both faces share similar eye shape and spacing, nose structure, and jawline contour. Eyebrows are moderately thick and arched similarly. Glasses are worn in both, though frames differ slightly. Overall facial structure and proportions align closely, supporting a high degree of similarity despite minor variations in lighting and image quality.}
]{\includegraphics[width=0.47\textwidth, alt=Image pair of two similari individuals with Qwen's justification.]{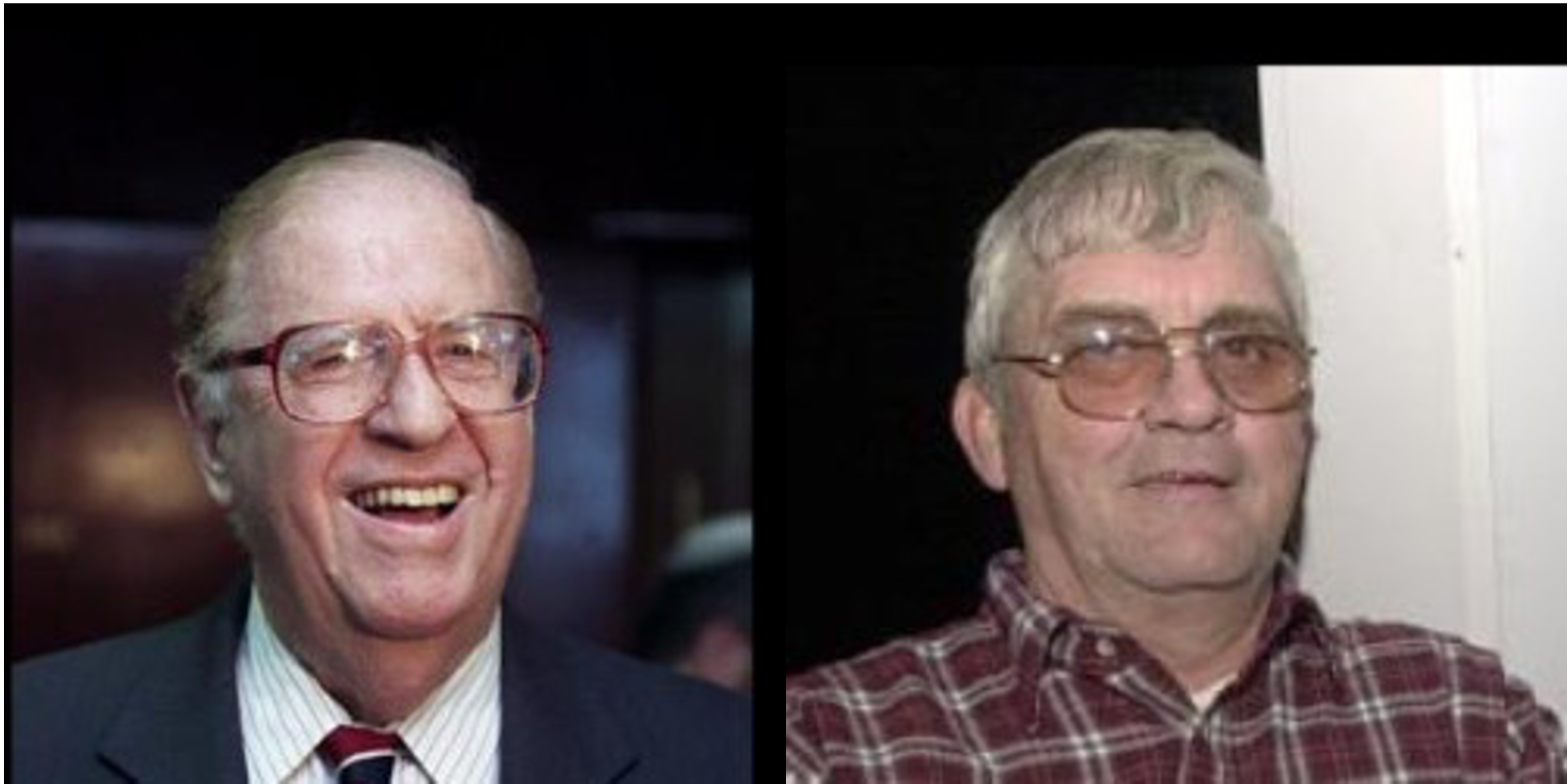}}
  \caption{Edge cases selected for analysing the quality of the explanations.}
  \label{fig:standalone-examples}
\end{figure*}

To further examine the explanations generated by the VLMs, we qualitatively analyzed a subset of challenging face pairs. Rather than evaluating the explanations against a predefined ground truth, which is not available for the visual evidence underlying an identity decision, we selected representative cases to illustrate the reasoning provided by the models. In particular, we considered pairs that were either close to the decision threshold despite being correctly classified, or far from the threshold despite being incorrectly classified. These cases allow us to examine how the models justify decisions under ambiguity and how their explanations change when multiple VLMs are fused. Figure \ref{fig:standalone-examples} shows representative face pairs from this analysis.

For image pair (a), the models consistently assigned a high similarity score, with Gemma producing the highest score (0.92), the explanation is grounded in the visual content of the images, refers to specific facial attributes, and identifies features that are relevant for distinguishing the two faces. Furthermore, the justification is consistent with the high similarity score assigned by the model and does not rely on information outside the provided images. This example therefore illustrates an important distinction between recognition accuracy and explanation quality: the model's decision is incorrect, while its justification is specific, visually grounded, and coherent with its prediction.

For image pair (b), Intern and Ovis both identify differences between the two faces in their justifications, including differences in perceived gender and facial characteristics, leading to an incorrect decision. In contrast, Gemma and Qwen correctly identify the two images as depicting the same individual. The incorrect decisions appear to be influenced by the presence of a second face in the first image and the pronounced pose of the target face, which may introduce misleading visual cues. This example highlights the value of textual explanations as a complementary layer of information for the operator. Even when the recognition decision is incorrect, the justification can expose the visual cues underlying the error, potentially facilitating the identification and analysis of system failures.

The third pair is a confident false match example, all models gave a high similarity score, like Ovis, Intern mentioned differences in hair and expression which are not related to facial attributes. This example illustrates a distinction between explanation correctness and verification correctness: the models correctly identify similarities between the two faces, but these similarities are not sufficiently discriminative to distinguish between different identities. Consequently, the explanations provide a plausible rationale for the models' decisions without establishing that the two images belong to the same person. This highlights an important limitation of explanation-based FR, where a visually accurate description of shared facial characteristics can nevertheless support an incorrect identity decision.

Finally, in the fourth example, despite Qwen identifying several similarities between the two faces, it assigns a relatively conservative similarity score of 0.75, comparable to those produced by Ovis and Intern. Gemma assigns an even lower score of 0.45 and highlights differences in the shape of the nose and chin. This example demonstrates that the models can adjust their similarity scores according to the degree of perceived similarity and provide explanations that are consistent with their final decisions.

\subsubsection{After Fusion Analysis}

After analysing the standalone explanations, we examined how the decider model combined the information provided by the source models. In scenario A, where the source models provided relatively consistent similarity estimates, the final score was generally close to their average, with a mean absolute difference between the mean of both source model similarity scores and the decider model's similarity score of 0.033. Suggesting that the decider model primarily aggregated the numerical evidence provided by the source models.

A different behaviour was observed in scenario B, when the source models provided substantially different similarity scores and, consequently, contradictory assessments of the image pair. Across the analysed cases, the decider model frequently tended to favour the more conservative assessment when the source models disagreed. In particular, when one model assigned a very high similarity score and another assigned a very low score, the decider model more frequently produced a score closer to the lower value. For example, when the source models assigned scores of 0.95 and 0.00, respectively, the decider model most often favoured the latter assessment after considering the accompanying justifications. This behaviour suggests that the decider model does not simply average the numerical scores when the source models disagree. Instead, the textual justifications appear to influence the final assessment, with the decider model potentially using the explanations to resolve disagreements between the source models. The tendency toward the lower score can be interpreted as a more conservative decision strategy; however, it should not necessarily be considered an improvement in recognition performance, since the lower-scoring source model may itself be incorrect. Nevertheless, this behaviour is relevant to the qualitative analysis because it demonstrates that the explanations provide information that can influence the fusion process beyond the numerical similarity scores alone.

\begin{figure*}
  \centering
    \subfigure[Ovis $\leftarrow$ Intern + Gemma: \textit{The first image shows two individuals with distinct facial features, including prominent cheekbones, eye shape, and jawlines, indicating high similarity. The second image depicts three men with varied facial structures, suggesting low similarity. Decision A's score of 0.0 aligns with the low similarity observed in the second image, while Decision B's score of 0.1 is slightly higher but still reflects minimal similarity. My own assessment of 0.9 for the first image contrasts sharply with the low scores provided, emphasizing the significant difference in facial features between the two sets of images. Fusing the scores results in an average of 0.33, acknowledging the disparity between the two scenarios.}
]{\includegraphics[width=0.47\textwidth, alt= Image pair of two individual with Ovis fused justification.]{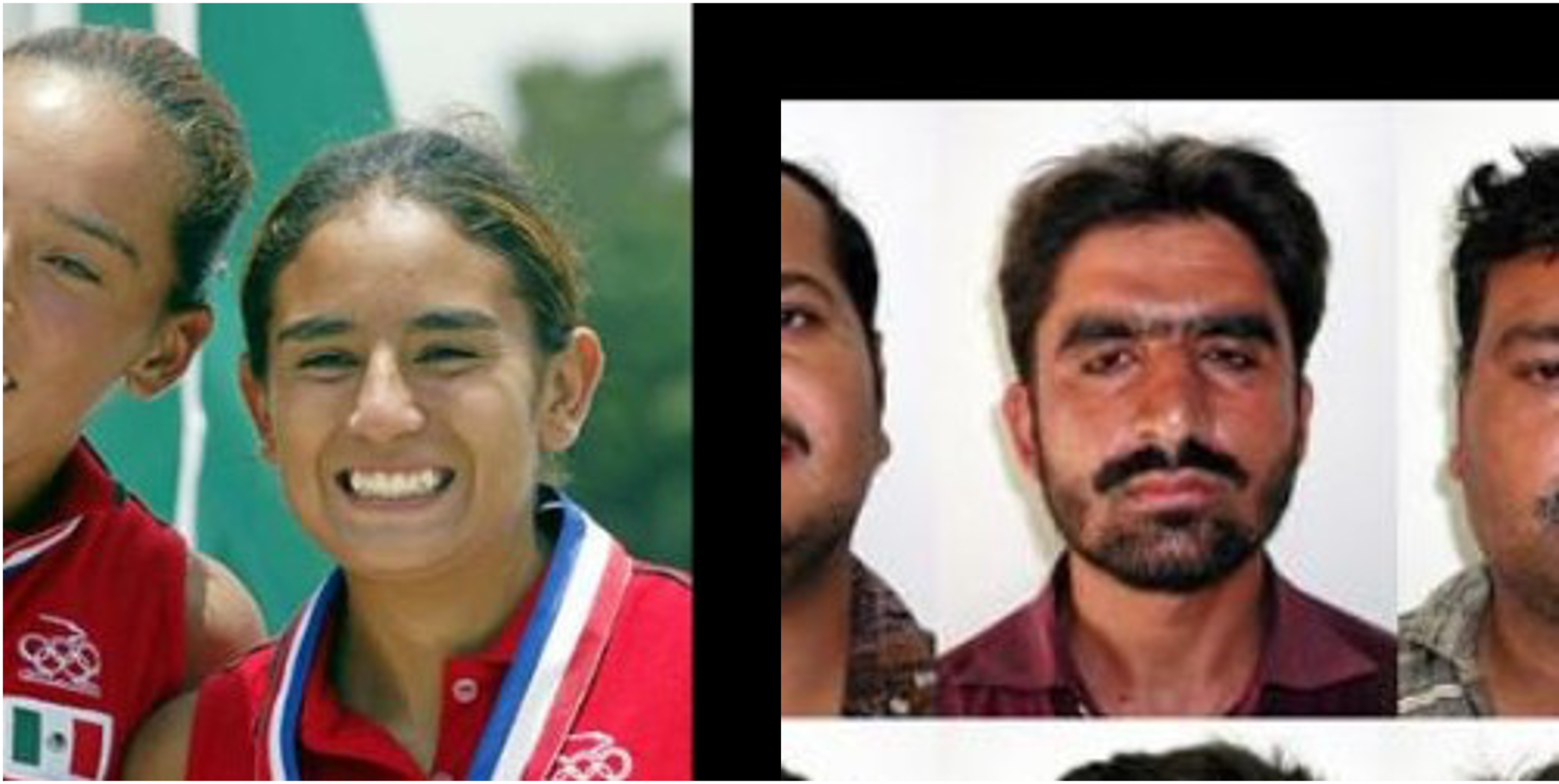}} \hfill
    \subfigure[Gemma $\leftarrow$ Qwen + Intern: \textit{My own assessment identifies a high degree of similarity between the central figure in both images, noting consistent eye shape, nose structure, and overall facial proportions. The prior results are highly inaccurate as they appear to compare the wrong individuals (e.g., comparing the man on the left in image 2 to the man in image 1, or the man on the right in image 1 to the man in image 2), rather than the primary subject present in both. Decision A and B both incorrectly conclude low similarity by focusing on different people within the frames. While my own score is high, the fused score is lowered to reflect the conflict with the prior decisions, although those decisions are fundamentally flawed in their subject selection.}
]{\includegraphics[width=0.47\textwidth, alt=image pair of two similari individuals with Gemma's fused justification.]{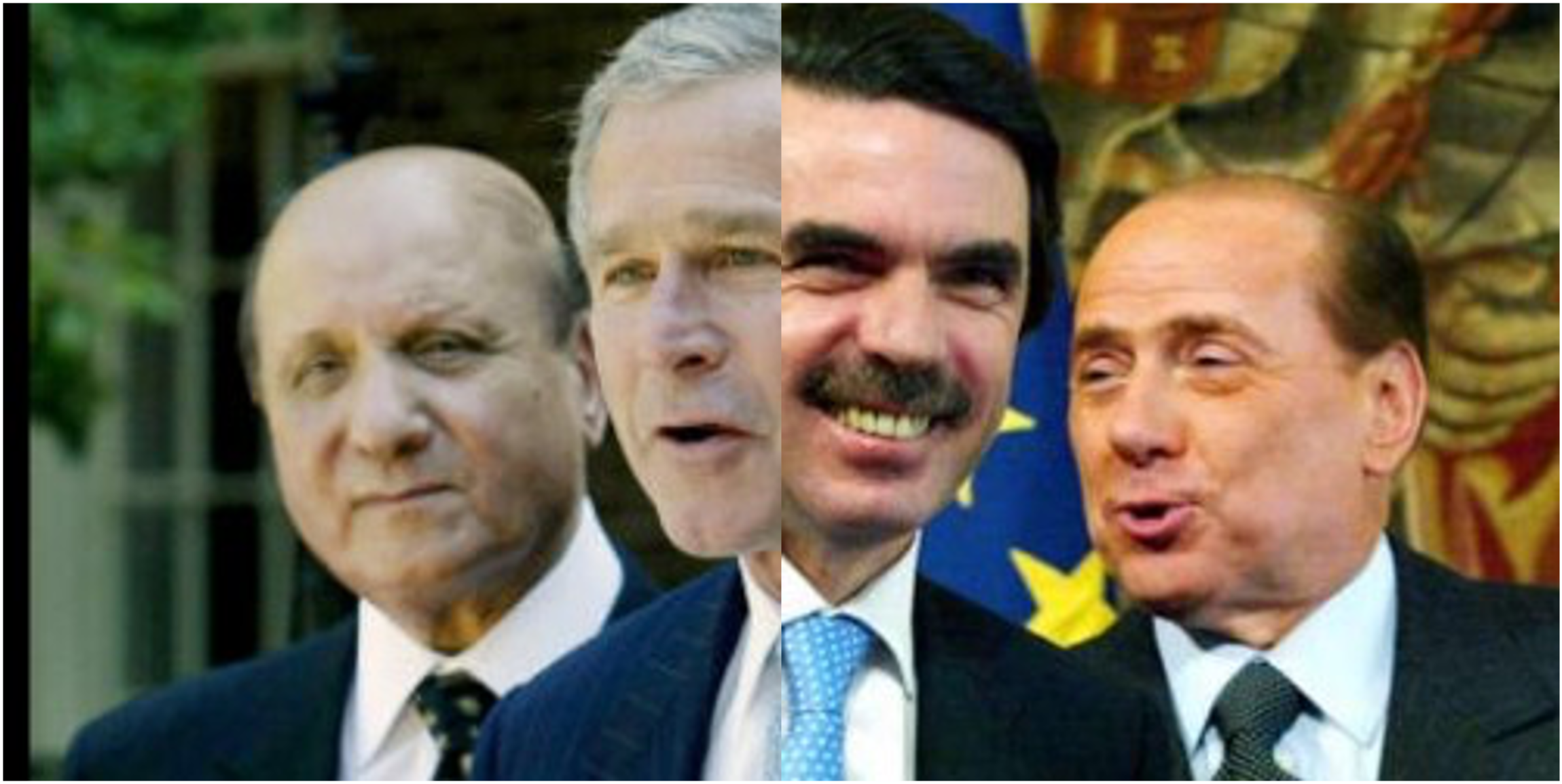}} 
  \caption{Edge cases selected for analysing the quality of the explanations, (a) corresponds to scenario C where Ovis had access to only the scores produced by Intern and Gemma, and (b) corresponds to scenario D where Gemma has access to the scores and justifications from Quen and Intern.}
  \label{fig:fusion-examples}
\end{figure*}

For scenario C, as the decider model has access to the images the possibility to override the source decisions arise, in the example shown in Figure \ref{fig:fusion-examples} (a) it is possible to observe that the decider model (Ovis) overrides the two similarity scores provided by the source models, the final similarity score provided corresponds to the two faces present in the first image. Giving the VLM access to the raw images increases its ability to resolve disagreements, but also exposes the fusion process to errors in visual grounding.

Regarding the results of scenario D, as shown in Figure \ref{fig:fusion-examples} (b), the decider model is able to interpret the reasoning behind the similarity scores provided by the source models: a difficult pair of images is provided to the VLMs as many faces are present, while Gemma focuses on the correct pair of faces present in the images and overrides the decision provided by the two source models' ones. For difficult examples, this becomes particularly important. A source model may provide a score suggesting a strong match, but its justification can reveal that it is actually comparing the target face with the wrong person in the image. The third VLM can then use the image itself to identify this inconsistency and evaluate the source recommendation.

The explanations do not merely provide additional support for the source models' predictions; they can also expose the basis of erroneous predictions, enabling the fusion model to explicitly identify and reject misleading source information. Multi-face cases are a specific class. However, they demonstrate why explanations can be useful even when the numerical scores alone appear informative: a score tells the third VLM how strongly a source model believes there is a match, whereas the justification can reveal which face and which visual evidence produced that belief. In a crowded image, that distinction can be decisive.

Taken together, these observations highlight that explanations provide value not only as a means of making VLM decisions more interpretable to human observers, but also as an additional source of information for other VLMs. While score-only fusion largely reduces the source models' outputs to numerical evidence, the inclusion of justifications provides insight into the reasoning underlying those scores. This additional information can allow the fusion model to adopt a more conservative decision when the source models disagree, identify inconsistencies in their reasoning, and, particularly in challenging images, recognize when a source model based its decision on the visual cues. Thus, explanations can serve as a form of machine-readable evidence, enabling downstream VLMs to assess not only what another model decided, but also the basis for that decision.

\subsection{Comparison with the state-of-the-art}

The results show that VLMs can achieve strong FR performance, consistent with the improvements reported by Shahreza et al. \cite{Shahreza_FRLLM_ICASSP_2026}, with Qwen2 getting closer in performance to CharGPT (93.50\%) \cite{DeAndresTame_FR_Corr_2024}. According to our experiments Gemma standalone achieves an AUC of 0.9985 and an EER of 1.30\%, compared with an AUC of 0.9741 and an EER of 5.57\% for the strongest conventional baseline, MagFace. Qwen also achieves strong overall separation, with a d-prime of 7.69. At 0.01\% FMR, Gemma reaches an FNMR of 21.38\%, whereas MagFace remains close to 6\%. Thus, although VLMs provide strong discrimination over the overall score distribution, their performance is less stable at stringent operating points, this might be due to the very discrete values that standalone VLMs output as shown in Figure \ref{fig:histograms}.

The proposed fusion strategy provides an improvement beyond the standalone VLM results and distinguishes this work from previous benchmarks. In the best combination of scenario D with Gemma $\leftarrow$ Qwen + Intern, it reduces the EER from 1.30\% for standalone Gemma to 1.06\% when the decider model has access to both the source outputs and the original images.

\section{Discussion}
\label{sec:Discussion}

The fusion experiments indicate that combining VLM outputs can further improve recognition performance, although the improvement depends strongly on the selected models and the information provided to the decider model. In the score-based setting, the best configuration reaches an EER of 1.17\%, while providing the decider model with the original images, scores, and explanations reduces the EER further to 1.06\% Gemma benefits from access to the original images, whereas some configurations using Intern and Ovis perform better on the score only and score and justification scenarios. Therefore, depending on the resources available, if time and computational power are limited, providing only similarity scores of different sources can improve FR performance. However, the qualitative analysis provides evidence of the complementary role of explanations. Specially in edge cases (close to the decision threshold) the explanations can reveal which facial characteristics a source model used to support its prediction, allowing the decider model to identify inconsistencies that are not apparent from the numerical scores alone. In some cases, the decider model is able to recognize that a source model compared the wrong individuals and revise the resulting assessment. This suggests that explanations can function as an additional form of information for cross-model reasoning, rather than solely as a mechanism for presenting decisions to a human operator. 

Several limitations should therefore be considered when interpreting these results. First, a widely established dataset was used in these experiments to be able to compare obatined results with the state-of-the-art. In future studies larger and even more diverse datasets could be used, specially ones that are not available to the public as these might have been used for the models' training. Second, the qualitative analysis of explanations requires standardize metrics which remains an open research direction. Beyond the use of automated solutions, include LLM-as-a-Judge and semantic similarity metrics like BERTScore, work needs to be done with humans-in-the-loop that ensure hallucination prevention and mitigation. Third, the improvements achieved using multimodal fusion (face images, scores and justifications), need to be weighted against the cost of increased use of computational resources. For example, a single combination of scenario D may not be executed in real time on commodity hardware. However, in the considered fusion scenarios no additional data is needed, so when data capturing is not an viable solution for training new models, the proposed fusion might be. 

\section{Conclusions}
\label{sec:Conclusions}

This work investigated the use of VLM fusion to evaluate both FR performance and human-understanding of recognition decisions. The results show that, VLMs exhibit strong FR capabilities, getting closer in performance to dedicated FR systems. More importantly, the results demonstrate that VLM-based fusion can improve not only the recognition performance of standalone VLMs, it also provides richer and more informative explanations of their decisions that can inform other VLMs and most importantly humans responsible of auditing these systems.

In particular, the fusion configuration in which the decider VLM had access to the image pairs, as well as the similarity scores and justifications produced by the source models, achieved the best overall results. The diversity of the source models further contributed to the quality of the fusion, suggesting that complementary models' perspectives can provide valuable information to the decision-making process. When the decision-making VLM had access to the images, the information provided by the source models acted as supporting evidence enriching the information provided with the model's final decision. In contrast, when the images were unavailable, the similarity scores tended to promote an aggregated or average decision and the justifications could sometimes shift the final decision towards more conservative assessment. This highlights the potential of combining quantitative similarity scores with qualitative reasoning as a means of making FR decisions more understandable and auditable by humans.

Overall, these findings suggest that VLM fusion can serve as a promising approach towards more robust, transparent, and human-oriented FR systems. Rather than relying solely on a single similarity score or decision, the proposed approach enables multiple model perspectives and their associated justifications to contribute to the final decision and its explanation. The investigation of larger and more diverse source model ensembles, a standardized methodology for evaluation of explanations, and evaluation protocols that more directly measure how much standalone and after-fusion explanations to improve human understanding and trust in final FR decisions are subject to future work.

\bibliographystyle{plainnat}
\bibliography{references}

@String{Computing = "Computing" }

@String{Computer = "{IEEE} Computer" }

@String{Springer = "Springer-Verlag" }

@article{Bai_Qwen_arXiv_2025,
  title   = {Qwen3-VL Technical Report},
  author  = {Bai, S. and Cai, Y. and Chen, R. and et al.},
  year    = {2025},
  journal = {arXiv preprint arXiv:2511.21631},
  doi     = {10.48550/arXiv.2511.21631}
}

@article{Lu_Ovis_arXiv_2025,
  title   = {Ovis2.5 Technical Report},
  author  = {Lu, S. and Li, Y. and Xia, Y. and et al.},
  year    = {2025},
  journal = {arXiv preprint arXiv:2508.11737}
}

@misc{Zhu_Intern_arXiv_2025,
  title         = {InternVL3: Exploring Advanced Training and Test-Time Recipes for Open-Source Multimodal Models},
  author        = {Zhu, J. and Wang, W. and Chen, Z. and et al.},
  year          = {2025},
  eprint        = {2504.10479},
  archivePrefix = {arXiv},
  primaryClass  = {cs.CV}
}

@misc{ElAbd_Gemma4_arXiv_2026,
  title         = {Gemma 4 Technical Report},
  author        = {S. El Abd and V. Aggarwal and R. Algayres and et al.},
  year          = {2026},
  eprint        = {2607.02770},
  archivePrefix = {arXiv},
  primaryClass  = {cs.CL}
}

@Article{Cascone_XFR_Elsevier_2023,
  author    = {Cascone, L. and Pero, C. and Proença, H.},
  journal   = {Image and Vision Computing},
  title     = {Visual and textual explainability for a biometric verification system based on piecewise facial attribute analysis},
  year      = {2023},
  issn      = {0262-8856},
  month     = Apr,
  pages     = {104645},
  volume    = {132},
  doi       = {10.1016/j.imavis.2023.104645},
  publisher = {Elsevier BV},
}

@Article{Benedict_LawFR_WLLaw_2022,
  author    = {T. J. Benedict},
  journal   = {Washington and Lee Law Review},
  title     = {The Computer Got It Wrong: Facial Recognition Technology and Establishing Probable Cause of Arrest},
  year      = {2022},
  number    = {849},
  pages     = {S225--S233},
  volume    = {79},
  url       = {https://scholarlycommons.law.wlu.edu/wlulr/vol79/iss2/7},
}

@INPROCEEDINGS{Sony_MLLMFR_IWBF_2026,
  author={Sony, R. and Jain, A. K. and Ross, A.},
  booktitle={2026 14th Intl. Workshop on Biometrics and Forensics (IWBF)}, 
  title={MLLM-Based Textual Explanations for Face Comparison}, 
  year={2026},
  volume={},
  number={},
  pages={1-6},
  doi={10.1109/IWBF68042.2026.11558159}}

@Article{Hannan_XFRNL_CoRR_2026,
  author        = {S. A. Hannan and H. T. Bukhari and T. Cantalapiedra and E. Ansar and M. Baali and R. Singh and B. Raj},
  journal       = {CoRR},
  title         = {VerLM: Explaining Face Verification Using Natural Language},
  year          = {2026},
  volume        = {abs/2601.01798},
  archiveprefix = {arXiv},
  bibsource     = {dblp computer science bibliography, https://dblp.org},
  doi           = {10.48550/ARXIV.2601.01798},
  eprint        = {2601.01798},
}

@Article{Doh_XFR_CoRR_2024,
  author        = {M. Doh and C. M. Rodrigues and N. Boutry and L. Najman and M. Mancas and H. Bersini},
  journal       = {CoRR},
  title         = {Bridging Human Concepts and Computer Vision for Explainable Face Verification},
  year          = {2024},
  volume        = {abs/2403.08789},
  archiveprefix = {arXiv},
  bibsource     = {dblp computer science bibliography, https://dblp.org},
  doi           = {10.48550/ARXIV.2403.08789},
  eprint        = {2403.08789},
}

@Article{Shadman_XFR_CoRR_2025,
  author        = {R. Shadman and D. Hou and F. Hussain and M. G. S. Murshed},
  journal       = {CoRR},
  title         = {Explainable Face Recognition via Improved Localization},
  year          = {2025},
  volume        = {abs/2505.03837},
  archiveprefix = {arXiv},
  bibsource     = {dblp computer science bibliography, https://dblp.org},
  doi           = {10.48550/ARXIV.2505.03837},
  eprint        = {2505.03837},
}

@InProceedings{Lu_XFR_FG_2024,
  author          = {Y. Lu and Z. Xu and T. Ebrahimi},
  title           = {Explainable Face Verification via Feature-Guided Gradient Backpropagation},
  year            = {2024},
  address         = {Istanbul, Turkiye},
  pages           = {1--5},
  publisher       = {IEEE},
  date            = {27-31 May 2024},
  doi             = {10.1109/FG59268.2024.10581925},
  eventdate       = {27-31 May 2024},
  eventtitleaddon = {Istanbul, Turkiye},
  isbn            = {979-8-3503-9495-5},
  issn            = {2326-5396},
  booktitle         = {2024 IEEE 18th Intl. Conf. on Automatic Face and Gesture Recognition (FG)},
}

@misc{EUAIAct_Article14_2024,
  author       = {{European Parliament and Council of the European Union}},
  title        = {{Regulation (EU) 2024/1689: Artificial Intelligence Act, Article 14 -- Human Oversight}},
  year         = {2024},
  month        = {6},
  day          = {13},
  howpublished = {Official Journal of the European Union},
  url          = {https://eur-lex.europa.eu/eli/reg/2024/1689/oj},
  note         = {Article 14},
}

@misc{EUAIAct_Article13_2024,
  author       = {{European Parliament and Council of the European Union}},
  title        = {{Regulation (EU) 2024/1689: Artificial Intelligence Act, Article 13 -- Transparency and Provision of Information to Deployers}},
  year         = {2024},
  month        = {6},
  day          = {13},
  howpublished = {Official Journal of the European Union},
  url          = {https://eur-lex.europa.eu/eli/reg/2024/1689/oj},
  note         = {Article 13},
}

@Article{Adebayo_XAI_CoRR_2018,
  author        = {J. Adebayo and J. Gilmer and M. Muelly and I. J. Goodfellow and M. Hardt and B. Kim},
  journal       = {CoRR},
  title         = {Sanity Checks for Saliency Maps},
  year          = {2018},
  volume        = {abs/1810.03292},
  archiveprefix = {arXiv},
  bibsource     = {dblp computer science bibliography, https://dblp.org},
  doi           = {10.48550/arxiv.1810.03292},
  eprint        = {1810.03292},
  url           = {http://arxiv.org/abs/1810.03292},
}

@Article{Rudin_XAI_NATMI_2019,
  author    = {C. Rudin},
  journal   = {Nat. Mach. Intell.},
  title     = {Stop explaining black box machine learning models for high stakes decisions and use interpretable models instead},
  year      = {2019},
  number    = {5},
  pages     = {206--215},
  volume    = {1},
  bibsource = {dblp computer science bibliography, https://dblp.org},
  doi       = {10.1038/S42256-019-0048-X},
}

@article{DoshiVelez_TowardsAR_arXiv_2017,
  title={Towards A Rigorous Science of Interpretable Machine Learning},
  author={F. Doshi-Velez and B. Kim},
  journal={arXiv: Machine Learning},
  year={2017},
  url={https://api.semanticscholar.org/CorpusID:11319376}
}

@Article{Zhou_DL_CoRR_2015,
  author        = {B. Zhou and A. Khosla and {\`{A}}. Lapedriza and A. Oliva and A. Torralba},
  journal       = {CoRR},
  title         = {Learning Deep Features for Discriminative Localization},
  year          = {2015},
  volume        = {abs/1512.04150},
  archiveprefix = {arXiv},
  bibsource     = {dblp computer science bibliography, https://dblp.org},
  doi           = {10.48550/arxiv.1512.04150},
  eprint        = {1512.04150},
  url           = {http://arxiv.org/abs/1512.04150},
}

@InProceedings{Selvaraju_GradCAM_ICCV_2017,
  author          = {R. R. Selvaraju and M. Cogswell and A. Das and R. Vedantam and D. Parikh and D. Batra},
  title           = {Grad-CAM: Visual Explanations from Deep Networks via Gradient-Based Localization},
  year            = {2017},
  address         = {Venice, Italy},
  pages           = {618--626},
  publisher       = {IEEE},
  date            = {22-29 Oct. 2017},
  doi             = {10.1109/ICCV.2017.74},
  eventdate       = {22-29 Oct. 2017},
  eventtitleaddon = {Venice, Italy},
  isbn            = {978-1-5386-1033-6},
  issn            = {2380-7504},
  booktitle         = {2017 IEEE Intl. Conf. on Computer Vision (ICCV)},
}

@Article{Lundberg_InterpretableAI_CoRR_2017,
  author        = {S. M. Lundberg and S. Lee},
  journal       = {CoRR},
  title         = {A unified approach to interpreting model predictions},
  year          = {2017},
  volume        = {abs/1705.07874},
  archiveprefix = {arXiv},
  bibsource     = {dblp computer science bibliography, https://dblp.org},
  doi           = {10.48550/arxiv.1705.07874},
  eprint        = {1705.07874},
  url           = {http://arxiv.org/abs/1705.07874},
}

@Article{Ribeiro_AI_CoRR_2016,
  author        = {M. T. Ribeiro and S. Singh and C. Guestrin},
  journal       = {CoRR},
  title         = {"Why Should {I} Trust You?": Explaining the Predictions of Any Classifier},
  year          = {2016},
  volume        = {abs/1602.04938},
  archiveprefix = {arXiv},
  bibsource     = {dblp computer science bibliography, https://dblp.org},
  doi           = {10.48550/arxiv.1602.04938},
  eprint        = {1602.04938},
  url           = {http://arxiv.org/abs/1602.04938},
}

@InProceedings{Buolamwini_GenderShades_PMLC_2018,
  author    = {J. Buolamwini and T. Gebru},
  booktitle = {Conf. on Fairness, Accountability and Transparency, {FAT} 2018, 23-24 February 2018, New York, NY, {USA}},
  title     = {Gender Shades: Intersectional Accuracy Disparities in Commercial Gender Classification},
  year      = {2018},
  editor    = {Sorelle A. Friedler and Christo Wilson},
  pages     = {77--91},
  publisher = {{PMLR}},
  series    = {Proceedings of Machine Learning Research},
  volume    = {81},
  bibsource = {dblp computer science bibliography, https://dblp.org},
  url       = {http://proceedings.mlr.press/v81/buolamwini18a.html},
}

@Article{OToole_FRDemographcis_IVC_2012,
  author    = {A. J. O'Toole and P. J. Phillips and X. An and J. P. Dunlop},
  journal   = {Image Vis. Comput.},
  title     = {Demographic effects on estimates of automatic face recognition performance},
  year      = {2012},
  number    = {3},
  pages     = {169--176},
  volume    = {30},
  bibsource = {dblp computer science bibliography, https://dblp.org},
  doi       = {10.1016/J.IMAVIS.2011.12.007},
}

@InProceedings{Suresh_ML_ACM_2021,
  author    = {H. Suresh and J. V. Guttag},
  booktitle = {{EAAMO} 2021: {ACM} Conf. on Equity and Access in Algorithms, Mechanisms, and Optimization, Virtual Event, USA, October 5 - 9, 2021},
  title     = {A Framework for Understanding Sources of Harm throughout the Machine Learning Life Cycle},
  year      = {2021},
  pages     = {17:1--17:9},
  publisher = {{ACM}},
  bibsource = {dblp computer science bibliography, https://dblp.org},
  doi       = {10.1145/3465416.3483305},
}

@Article{Gebru_Data_ACM_2021,
  author    = {T. Gebru and J. Morgenstern and B. Vecchione and J. W. Vaughan and H. M. Wallach and H. Daum{\'{e}} III and K. Crawford},
  journal   = {Commun. {ACM}},
  title     = {Datasheets for datasets},
  year      = {2021},
  number    = {12},
  pages     = {86--92},
  volume    = {64},
  bibsource = {dblp computer science bibliography, https://dblp.org},
  doi       = {10.1145/3458723},
}

@Article{Baker_XAI_Corr_2023,
  author        = {S. Baker and W. Xiang},
  journal       = {CoRR},
  title         = {Explainable {AI} is Responsible {AI:} How Explainability Creates Trustworthy and Socially Responsible Artificial Intelligence},
  year          = {2023},
  volume        = {abs/2312.01555},
  archiveprefix = {arXiv},
  bibsource     = {dblp computer science bibliography, https://dblp.org},
  doi           = {10.48550/ARXIV.2312.01555},
  eprint        = {2312.01555},
}

@Book{Crawford_AIEthics_YalePress_2021,
  author    = {Crawford, K.},
  publisher = {Yale University Press},
  title     = {Atlas of AI},
  year      = {2021},
  address   = {New Haven},
  isbn      = {9780300252392},
  pagetotal = {1327},
  ppn_gvk   = {1752921879},
  subtitle  = {Power, politics, and the planetary costs of artificial intelligence},
}

@Article{Mehrabi_Fairness_ACMC_2021,
  author    = {Mehrabi, N. and Morstatter, F. and Saxena, N. and Lerman, K. and Galstyan, A.},
  journal   = {ACM Computing Surveys},
  title     = {A Survey on Bias and Fairness in Machine Learning},
  year      = {2021},
  issn      = {1557-7341},
  month     = {July},
  number    = {6},
  pages     = {1--35},
  volume    = {54},
  doi       = {10.1145/3457607},
  publisher = {Association for Computing Machinery (ACM)},
}

@Article{Morley_AIEthics_SEE_2019,
  author    = {Morley, J. and Floridi, L. and Kinsey, L. and Elhalal, A.},
  journal   = {Science and Engineering Ethics},
  title     = {From What to How: An Initial Review of Publicly Available AI Ethics Tools, Methods and Research to Translate Principles into Practices},
  year      = {2019},
  issn      = {1471-5546},
  month     = Dec,
  number    = {4},
  pages     = {2141--2168},
  volume    = {26},
  doi       = {10.1007/s11948-019-00165-5},
  publisher = {Springer Science and Business Media LLC},
}

@Article{Jobin_AIEthics_NMI_2019,
  author    = {Jobin, A. and Ienca, M. and Vayena, E.},
  journal   = {Nature Machine Intelligence},
  title     = {The global landscape of AI ethics guidelines},
  year      = {2019},
  issn      = {2522-5839},
  month     = {September},
  number    = {9},
  pages     = {389--399},
  volume    = {1},
  doi       = {10.1038/s42256-019-0088-2},
  publisher = {Springer Science and Business Media LLC},
}

@Article{Floridi_AIEthics_MaM_2018,
  author    = {L. Floridi and J. Cowls and M. Beltrametti and R. Chatila and P. Chazerand and V. Dignum and C. Luetge and R. Madelin and U. Pagallo and F. Rossi and B. Schafer and P. Valcke and E. Vayena},
  journal   = {Minds Mach.},
  title     = {AI4People - An Ethical Framework for a Good {AI} Society: Opportunities, Risks, Principles, and Recommendations},
  year      = {2018},
  number    = {4},
  pages     = {689--707},
  volume    = {28},
  bibsource = {dblp computer science bibliography, https://dblp.org},
  doi       = {10.1007/S11023-018-9482-5},
}

@InProceedings{Wang_FR_CVPR_2018,
author = {Wang, H. and Wang, Y. and Zhou, Z. and Ji, X. and Gong, D. and Zhou, J. and Li, Z. and Liu, W.},
title = {CosFace: Large Margin Cosine Loss for Deep Face Recognition},
booktitle = {Proceedings of the IEEE Conf. on Computer Vision and Pattern Recognition (CVPR)},
month = {June},
year = {2018}
}

@InProceedings{Liu_FR_CVPR_1017,
author = {Liu, W. and Wen, Y. and Yu, Z. and Li, M. and Raj, B. and Song, L.},
title = {SphereFace: Deep Hypersphere Embedding for Face Recognition},
booktitle = {Proceedings of the IEEE Conf. on Computer Vision and Pattern Recognition (CVPR)},
month = {July},
year = {2017}
}

@InProceedings{You_LVFace_ICCV_2025,
  author          = {J. You and S. Li and Y. Sun and J. Wei and M. Guo and C. Feng and J. Ran},
  title           = {LVFace: Progressive Cluster Optimization for Large Vision Models in Face Recognition},
  year            = {2025},
  address         = {Honolulu, HI, USA},
  pages           = {11840--11849},
  publisher       = {IEEE},
  date            = {19-25 Oct. 2025},
  doi             = {10.1109/ICCV51701.2025.01101},
  eventdate       = {19-25 Oct. 2025},
  eventtitleaddon = {Honolulu, HI, USA},
  isbn            = {979-8-3315-8776-5},
  issn            = {1550-5499},
  booktitle         = {2025 IEEE/CVF Intl. Conf. on Computer Vision (ICCV)},
}

@InProceedings{Meng_MagFace_CVPR_2021,
  author          = {Q. Meng and S. Zhao and Z. Huang and F. Zhou},
  title           = {MagFace: A Universal Representation for Face Recognition and Quality Assessment},
  year            = {2021},
  address         = {Nashville, TN, USA},
  pages           = {14220--14229},
  publisher       = {IEEE},
  date            = {20-25 June 2021},
  doi             = {10.1109/CVPR46437.2021.01400},
  eventdate       = {20-25 June 2021},
  eventtitleaddon = {Nashville, TN, USA},
  isbn            = {978-1-6654-4510-8},
  issn            = {1063-6919},
  booktitle         = {2021 IEEE/CVF Conf. on Computer Vision and Pattern Recognition (CVPR)},
}

@InProceedings{Kim_AdaFace_CVPR_2022,
  author          = {M. Kim and A. K. Jain and X. Liu},
  title           = {AdaFace: Quality Adaptive Margin for Face Recognition},
  year            = {2022},
  address         = {New Orleans, LA, USA},
  pages           = {18729--18738},
  publisher       = {IEEE},
  date            = {18-24 June 2022},
  doi             = {10.1109/CVPR52688.2022.01819},
  eventdate       = {18-24 June 2022},
  eventtitleaddon = {New Orleans, LA, USA},
  isbn            = {978-1-6654-6947-0},
  issn            = {1063-6919},
  booktitle         = {2022 IEEE/CVF Conf. on Computer Vision and Pattern Recognition (CVPR)},
}

@TechReport{Huang_LFWTech_2007,
    author = {G. B. Huang and M. Ramesh and T. Berg and E. Learned-Miller},
    title = {Labeled Faces in the Wild: A Database for Studying Face Recognition in Unconstrained Environments},
    institution = {University of Massachusetts, Amherst},
    year = 2007,
    number = {07-49},
    month = {October}
}

@InProceedings{Huber_XFR_WACV_2025,
  author          = {M. Huber and N. Damer},
  title           = {Beyond Spatial Explanations: Explainable Face Recognition in the Frequency Domain},
  year            = {2025},
  address         = {Tucson, AZ, USA},
  pages           = {1016--1026},
  publisher       = {IEEE},
  date            = {26 Feb.-6 March 2025},
  doi             = {10.1109/WACV61041.2025.00108},
  eventdate       = {26 Feb.-6 March 2025},
  eventtitleaddon = {Tucson, AZ, USA},
  isbn            = {979-8-3315-1084-8},
  issn            = {2472-6737},
  booktitle         = {2025 IEEE/CVF Winter Conf. on Applications of Computer Vision (WACV)},
}

@Article{DeAndresTame_XFR_Corr_2024a,
  author        = {I. DeAndres{-}Tame and M. Faisal and R. Tolosana and R. Al{-}Refai and R. Vera{-}Rodr{\'{\i}}guez and P. Terh{\"{o}}rst},
  journal       = {CoRR},
  title         = {From Pixels to Words: Leveraging Explainability in Face Recognition through Interactive Natural Language Processing},
  year          = {2024},
  volume        = {abs/2409.16089},
  archiveprefix = {arXiv},
  bibsource     = {dblp computer science bibliography, https://dblp.org},
  doi           = {10.48550/ARXIV.2409.16089},
  eprint        = {2409.16089},
}

@InProceedings{Lu_XFR_WACV_2024,
  author          = {Y. Lu and Z. Xu and T. Ebrahimi},
  title           = {Towards Visual Saliency Explanations of Face Verification},
  year            = {2024},
  address         = {Waikoloa, HI, USA},
  pages           = {4714--4723},
  publisher       = {IEEE},
  date            = {3-8 Jan. 2024},
  doi             = {10.1109/WACV57701.2024.00466},
  eventdate       = {3-8 Jan. 2024},
  eventtitleaddon = {Waikoloa, HI, USA},
  isbn            = {979-8-3503-1893-7},
  issn            = {2472-6737},
  booktitle         = {2024 IEEE/CVF Winter Conf. on Applications of Computer Vision (WACV)},
}

@Article{DeAndresTame_FR_Corr_2024,
  author        = {I. DeAndres{-}Tame and R. Tolosana and R. Vera{-}Rodr{\'{\i}}guez and A. Morales and J. Fi{\'{e}}rrez and J. Ortega{-}Garcia},
  journal       = {CoRR},
  title         = {How Good is ChatGPT at Face Biometrics? {A} First Look into Recognition, Soft Biometrics, and Explainability},
  year          = {2024},
  volume        = {abs/2401.13641},
  archiveprefix = {arXiv},
  bibsource     = {dblp computer science bibliography, https://dblp.org},
  doi           = {10.48550/ARXIV.2401.13641},
  eprint        = {2401.13641},
}

@misc{Grother_Identification_NIST_2019,
  author = {P. Grother and M. Ngan and K. Hanaoka},
  title = {Face Recognition Vendor Test (FRVT) Part 2: Identification},
  year = {2019},
  month = {2019-09-13 00:09:00},
  publisher = {NIST Interagency/Internal Report (NISTIR), National Institute of Standards and Technology, Gaithersburg, MD},
  doi = {https://doi.org/10.6028/NIST.IR.8271},
  language = {en},
}

@misc{Grother_FR_NIST_2019,
  author = {P. Grother and M. Ngan and K. Hanaoka},
  title = {Face Recognition Vendor Test Part 3: Demographic Effects},
  year = {2019},
  month = {2019-12-19 00:12:00},
  publisher = {NIST Interagency/Internal Report (NISTIR), National Institute of Standards and Technology, Gaithersburg, MD},
  doi = {https://doi.org/10.6028/NIST.IR.8280},
  language = {en},
}

@InProceedings{Deng_ArcFace_CVPR_2019,
  author          = {J. Deng and J. Guo and N. Xue and S. Zafeiriou},
  title           = {ArcFace: Additive Angular Margin Loss for Deep Face Recognition},
  year            = {2019},
  address         = {Long Beach, CA, USA},
  pages           = {4685--4694},
  publisher       = {IEEE},
  date            = {15-20 June 2019},
  doi             = {10.1109/CVPR.2019.00482},
  eventdate       = {15-20 June 2019},
  eventtitleaddon = {Long Beach, CA, USA},
  isbn            = {978-1-7281-3294-5},
  issn            = {1063-6919},
  booktitle         = {2019 IEEE/CVF Conf. on Computer Vision and Pattern Recognition (CVPR)}
}

@inproceedings{Schroff-FaceNet-CVPR-2015,
 Author = {F. Schroff and D. Kalenichenko and J. Philbin},
 Booktitle = {Proc. {IEEE} Conf. on Computer Vision and Pattern Recognition ({CVPR})},
 Pages = {815--823},
 Title = {{FaceNet:} A Unified Embedding for Face Recognition and Clustering},
 Year = {2015}
}

@inproceedings{Taigman-DeepFace-CVPR-2014,
 Author = {Y. Taigman and M. Yang and M. Ranzato and L. Wolf},
 Booktitle = {2014 {IEEE} Conf. on Computer Vision and Pattern Recognition ({CVPR})},
 Month = {June},
 Pages = {1701--1708},
 Title = {{DeepFace}: Closing the Gap to Human-Level Performance in Face Verification},
 Year = {2014}
}

@misc{EU-Regulation-AI-2024-1689-en-240712,
 Author = {{European Council}},
 Month = {July},
 Title = {Regulation 2024/1689 of the European Parliament and of the Council of 13 June 2024 on laying down harmonised rules on artificial intelligence({AI-Act})},
 Year = {2024}
}

@INPROCEEDINGS{Shahreza_FULLM_ICCVW_2025,
  author={Shahreza, H. O. and Marcel, S.},
  booktitle={2025 IEEE/CVF Intl. Conf. on Computer Vision Workshops (ICCVW)}, 
  title={FaceLLM: A Multimodal Large Language Model for Face Understanding}, 
  year={2025},
  volume={},
  number={},
  pages={3736-3746},
  doi={10.1109/ICCVW69036.2025.00390}}

@ARTICLE{Narayan_FU_IEEETBIOM_2026,
  author={Narayan, K. and Vibashan, V. S. and Patel, V. M.},
  journal={IEEE Transactions on Biometrics, Behavior, and Identity Science}, 
  title={FaceXBench: Evaluating Multimodal LLMs on Face Understanding}, 
  year={2026},
  volume={8},
  number={3},
  pages={354-364},
  doi={10.1109/TBIOM.2026.3655668}}

@InProceedings{Sony_FRLLM_ICCVW_2025,
  author          = {R. Sony and P. Farmanifard and A. Ross and A. K. Jain},
  title           = {Foundation Versus Domain-Specific Models: Performance Comparison, Fusion, and Explainability in Face Recognition},
  year            = {2025},
  address         = {Honolulu, HI, USA},
  pages           = {3715--3725},
  publisher       = {IEEE},
  date            = {19-20 Oct. 2025},
  doi             = {10.1109/ICCVW69036.2025.00388},
  eventdate       = {19-20 Oct. 2025},
  eventtitleaddon = {Honolulu, HI, USA},
  isbn            = {979-8-3315-8989-9},
  issn            = {2473-9936},
  booktitle         = {2025 IEEE/CVF Intl. Conf. on Computer Vision Workshops (ICCVW)},
}

@InProceedings{Shahreza_FRLLM_ICASSP_2026,
  author          = {H. O. Shahreza and S. Marcel},
  title           = {Benchmarking Multimodal Large Language Models for Face Recognition},
  year            = {2026},
  address         = {Barcelona, Spain},
  pages           = {13557--13561},
  publisher       = {IEEE},
  date            = {3-8 May 2026},
  doi             = {10.1109/ICASSP55912.2026.11463782},
  eventdate       = {3-8 May 2026},
  eventtitleaddon = {Barcelona, Spain},
  isbn            = {979-8-3315-6702-6},
  issn            = {1520-6149},
  booktitle         = {ICASSP 2026 - 2026 IEEE Intl. Conf. on Acoustics, Speech and Signal Processing (ICASSP)},
}

\end{document}